\documentclass[final,5p,times,twocolumn,authoryear]{elsarticle}

\usepackage{amsmath,amssymb}
\usepackage{graphicx}
\usepackage{booktabs,multirow,array,tabularx,makecell}
\usepackage{pifont}
\usepackage[most]{tcolorbox}
\usepackage{url}
\usepackage[hidelinks]{hyperref}
\usepackage{orcidlink}

\newcommand{\ours}{RSure-Agent}
\newcommand{\base}{RSure-Base}
\newcommand{\cmark}{\ding{51}}
\newcommand{\xmark}{\ding{55}}
\newcommand{\authororcid}[1]{\,\orcidlink{#1}}
\newtcolorbox{promptbox}[1]{
  breakable,
  title={#1},
  colback=white,
  colframe=black,
  boxrule=0.5pt,
  arc=1pt,
  left=4pt,
  right=4pt,
  top=4pt,
  bottom=4pt,
  fonttitle=\bfseries
}

\biboptions{authoryear,round}
\journal{ISPRS Journal of Photogrammetry and Remote Sensing}

\begin{document}
\begin{frontmatter}

\title{RSure-Agent: Reliable Use of Tool Observations for Remote Sensing Agents}

\author[tju]{Fuyuan~Liu\authororcid{0009-0000-5776-6136}}
\ead{fuyuan695104062@tju.edu.cn}
\author[tju]{Nayu~Liu\authororcid{0000-0002-7664-9856}\corref{cor1}}
\ead{nyliu@tju.edu.cn}
\author[tju]{Wenhao~Yu\authororcid{0009-0006-1421-7050}}
\author[tcat]{Peijin~Wang\authororcid{0000-0002-0371-5584}}
\author[tcat]{Yingchao~Feng\authororcid{0000-0003-4017-8885}}
\author[tcat]{Fanglong~Yao\authororcid{0000-0003-4187-9755}}
\author[ecnu]{Yue~Zhou\authororcid{0000-0002-3080-6721}}
\author[sjtu]{Xue~Yang\authororcid{0000-0002-7084-9101}}
\author[tcat]{Chao~Ren\authororcid{0000-0001-9096-8792}}
\author[tju]{Liang~Wan\authororcid{0000-0001-5501-9575}}
\author[tju]{Wei~Feng\authororcid{0000-0003-3809-1086}}

\cortext[cor1]{Corresponding author: Nayu Liu.}

\affiliation[tju]{
  organization={School of Computer Science and Technology, Tianjin University},
  city={Tianjin},
  postcode={300350},
  country={China}
}

\affiliation[tcat]{
  organization={Key Laboratory of Target Cognition and Application Technology (TCAT),
  Aerospace Information Research Institute, Chinese Academy of Sciences},
  city={Beijing},
  postcode={100094},
  country={China}
}

\affiliation[ecnu]{
  organization={School of Geospatial Artificial Intelligence,
  East China Normal University},
  city={Shanghai},
  postcode={200241},
  country={China}
}

\affiliation[sjtu]{
  organization={School of Automation and Intelligent Sensing,
  Shanghai Jiao Tong University},
  city={Shanghai},
  postcode={200240},
  country={China}
}

%\begin{highlights}
%\item Quantifies tool observation errors and propagation across RS agent benchmarks.
%\item Proposes a verifiable observation protocol returning traceable process evidence.
%\item Builds a task--tool reliability prior from offline feedback to guide verification.
%\item Cuts error propagation by over 21 percentage points across three RS benchmarks.
%\end{highlights}

\begin{abstract}
Remote sensing agents rely on perception, measurement, and raster analysis tools to solve Earth observation tasks. We refer to their judgments and quantitative results about ground objects as tool observations. However, these observations are subject to substantial uncertainty and may be incorrect even when the tools execute successfully. When agents accept incorrect observations, the errors can propagate through subsequent reasoning and cause task failure. We analyze 1,229 execution trajectories across three remote sensing agent benchmarks. On each benchmark, at least 88.1\% of tasks depend on tool observations. Among these tasks, at least 22.7\% contain incorrect observations despite successful tool execution. These errors propagate to the final answer in at least 82.0\% of affected tasks on each benchmark. To address this problem, we propose \ours{}, a framework for verifying tool observations and limiting error propagation. We introduce a verifiable observation protocol that requires tools to return process evidence for the agent to verify their observations. We also construct a task--tool reliability prior from offline task feedback. The prior summarizes each tool configuration's past performance across task types and provides a task-specific reference for verification. Using process evidence and this prior, \ours{} decides whether to accept an observation, request additional evidence, or reject it. We evaluate \ours{} on EarthBench, ThinkGeo, TerraLogic, and CHOICE-420. Across the three agent benchmarks, \ours{} reduces the error propagation rate by 21.3 to 25.9 percentage points relative to the base configuration with verification and the prior disabled. On CHOICE-420, it improves overall accuracy over direct answering by 5.71 percentage points on average across 11 backbone models. On EarthBench, it reduces the tool-call ratio by 25.9\% relative to Earth-Agent. The demo is publicly available at \url{https://github.com/airs101/RSure-Agent}, and the source code will also be released for further research.
\end{abstract}

\begin{keyword}
Earth observation \sep Remote sensing agents \sep Observation verification
\sep Task--tool reliability prior
\end{keyword}

\end{frontmatter}

\begin{figure*}[t]
\centering
\includegraphics[width=\textwidth]{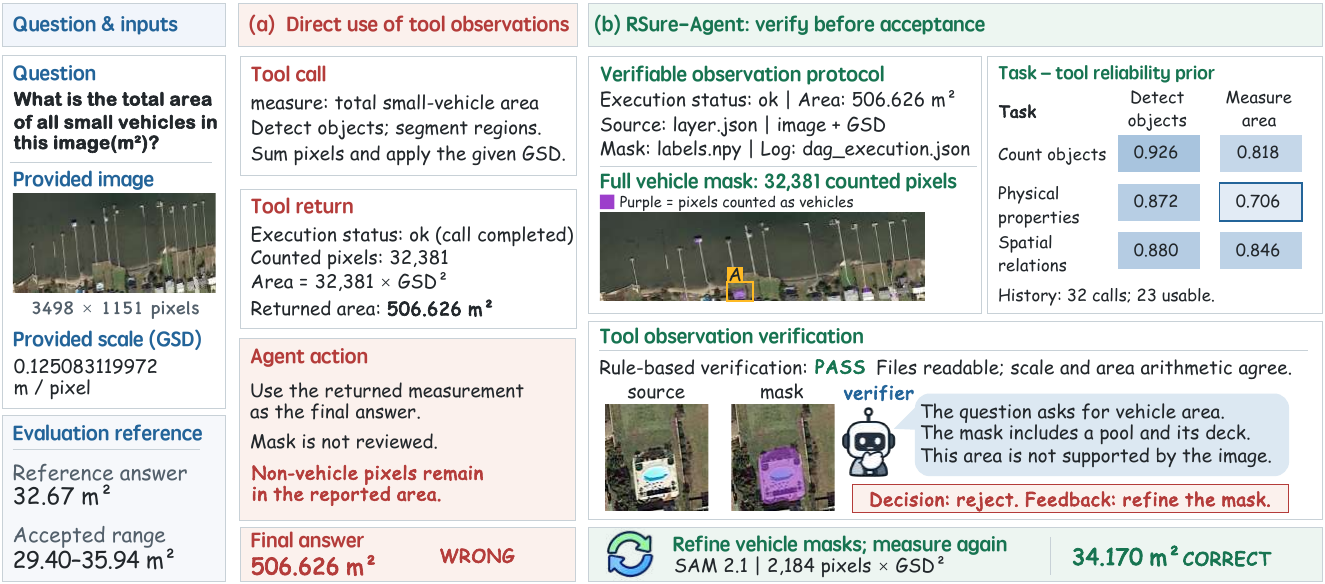}
\caption{Illustration of \ours{}. (a) The tool executes without runtime errors, but the segmentation mask includes non-vehicle pixels, yielding an erroneous area of $506.626~\mathrm{m}^{2}$. (b) Through the verifiable observation protocol, \ours{} inspects the source image, mask, and execution log. After deterministic checks pass, the verifier detects that the mask covers a swimming pool, refuses the faulty observation, and prompts mask refinement. Following SAM~2.1 refinement, the recomputed area is $34.170~\mathrm{m}^{2}$ (within the accepted range of $29.40$--$35.94~\mathrm{m}^{2}$). The task--tool reliability prior provides a historical reference during verification.}
\label{fig:motivation}
\end{figure*}

\section{Introduction}
\label{intro}

Earth observation is essential for understanding the dynamics of the Earth system. Remote sensing (RS) vision-language models link image content with natural language~\citep{hu2025rsgpt,zhan2025skyeyegpt}. These capabilities support the development of RS agents that automate Earth observation tasks through planning and external tool use~\citep{talemi2026agentic,xiao2026geomm}. Agents combine specialized tools according to task requirements. They also decide when to use their own capabilities and when to call external tools.  During execution, agents use visual perception, spatial measurement, and raster analysis tools to obtain judgments and quantitative results about ground objects. These results include object categories, geometric distances, areas, and remote sensing products. Following the definition of observations in Earth observation~\citep{iso19156,strobl2024lost}, we refer to these outputs as \textbf{tool observations}. The term differs from observation in agent paradigms such as ReAct~\citep{yao2023react}, where it refers to feedback from tool execution.

Existing remote sensing agents typically build on general-purpose agent frameworks~\citep{wang2024llmagentsurvey,xi2023llmagentrise}. Their methods combine task planning~\citep{feng2026earthagent,lee2025geollmsquad,dai2026geoevolver}, knowledge retrieval and memory~\citep{wu2026rsmem}, and tool calling~\citep{xu2026rsagent,guo2024remote} to solve complex tasks. For tool use, these methods focus mainly on correct selection and successful execution. Tool observations are usually passed directly to subsequent reasoning after successful execution~\citep{shabbir2026openearthagent,munir2026agenticrs}. However, successful execution does not guarantee reliable observations. Studies of semantic segmentation, object detection, hyperspectral classification, and target extraction link result reliability to scene complexity, image quality, and prediction uncertainty~\citep{rssegquality2025,saruncertainty2023,hsiuncertainty2022,uanet2024,li2026pugnet}. For example, in change detection, image misalignment, sensor differences, and seasonal variation can lead to false detections~\citep{jing2025changerd}. If the agent accepts an incorrect observation, later steps may rely on it. The error can then propagate and cause task failure.

We analyze 1,229 agent execution trajectories from three benchmarks to quantify this problem. These are EarthBench~\citep{feng2026earthagent}, ThinkGeo~\citep{shabbir2025thinkgeo}, and TerraLogic~\citep{yan2026terralogic}; Section~\ref{sec:empirical_analysis} presents the analysis. On each benchmark, at least 88.1\% of tasks depend on tool observations. Among these tasks, at least 22.7\% contain incorrect observations despite successful tool execution. On each benchmark, errors reach the final answer in at least 82.0\% of tasks containing incorrect observations. Tool observation reliability is therefore a recurring challenge for remote sensing agents.

To address this problem, we propose \ours{} to use tool observations more reliably and limit the impact of tool observation errors on subsequent reasoning. Fig.~\ref{fig:motivation} illustrates \ours{} in a task that measures the area of small vehicles. We design a verifiable observation protocol that requires each tool to return process evidence with its observation. Process evidence exposes intermediate results and computation conditions for verification. The verifier checks whether this evidence supports accepting the current observation. We build a task--tool reliability prior from offline trajectories. The prior records how different tool configurations perform on each task type. When the prior is low, the verifier checks the evidence more strictly. When the prior is high, the observation must still be supported by the image and process evidence. After a tool returns an observation, \ours{} combines these two sources to accept it, request additional evidence, or reject it. This reduces the effect of incorrect observations on subsequent reasoning.

We evaluate \ours{} on three remote sensing agent benchmarks: EarthBench, ThinkGeo, and TerraLogic~\citep{shabbir2025thinkgeo,feng2026earthagent,yan2026terralogic}. We also construct CHOICE-420 from CHOICE~\citep{li2025choice} to evaluate remote sensing visual question answering (VQA).
On agent benchmarks, \ours{} achieves competitive task performance. Relative to the base configuration with verification and the prior disabled, \ours{} reduces the error propagation rate by 21.3 to 25.9 percentage points. On CHOICE-420, it improves overall accuracy over direct answering by 5.71 percentage points on average across 11 backbone models.

Our main contributions are as follows:
\begin{itemize}
\item We quantify tool observation reliability across three RS agent benchmarks. On each benchmark, at least 88.1\% of tasks depend on tool observations. Among these tasks, at least 22.7\% contain incorrect tool observations despite successful tool execution. Among tasks with observation errors, the error propagation rate is at least 82.0\%.
\item We propose \ours{}, a framework limiting how incorrect tool observations affect subsequent reasoning. Its verifiable observation protocol requires process evidence from tools for verifying observations before acceptance.
\item We introduce a task--tool reliability prior to account for differences in tool reliability across task types. Derived from offline task feedback, the prior summarizes tool configuration performance and guides verification strictness.
\item We evaluate \ours{} on RS agent benchmarks and CHOICE-420. It reduces the error propagation rate by 21.3 to 25.9 percentage points on agent benchmarks and improves average VQA accuracy across 11 backbones.
\end{itemize}

\section{Related Work}
\label{sec:relatedwork}

\subsection{Remote Sensing Agents and Benchmarks}
\label{sec:rs_agent}

Remote sensing foundation models and VLMs provide visual perception and language understanding capabilities for Earth observation. Pre-trained visual and spectral foundation models learn representations across optical, multispectral, and multi-modal imagery~\citep{sun2022ringmo,guo2024skysense,hong2024spectralgpt}. Vision-language pre-training aligns remote sensing imagery with text for cross-modal retrieval and bi-temporal change analysis~\citep{liu2024remoteclip,dong2024changeclip,zavras2025mind}. Instruction-tuned VLMs further support image captioning~\citep{ge2025rsteller}, grounded visual question answering, and multi-sensor dialogue~\citep{hu2025rsgpt,kuckreja2024geochat,zhang2024earthgpt,zhan2025skyeyegpt,li2025lhrsbotnova}. In addition, reasoning-based segmentation models connect natural language queries with pixel-level target extraction~\citep{zhang2026think2segrs}. However, standalone foundation models and VLMs predict answers in a single forward pass and cannot directly execute calibrated geometric measurements or multi-step raster computations. To address these tasks, remote sensing agents combine foundation models with task planning and external tool use~\citep{talemi2026agentic,xiao2026geomm}. RS agents coordinate specialized tools for image perception, spatial analysis, and quantitative computation across multiple interaction steps~\citep{feng2026earthagent,xu2026rsagent,remoteagent2026}.

Early work explores how to combine remote sensing models with natural language interaction. Remote Sensing ChatGPT calls remote sensing models through dialogue~\citep{guo2024remote}. Tree-GPT enables interactive interpretation of forest remote sensing images~\citep{du2023tree}. Change-Agent combines change perception with language interaction to interpret land surface changes~\citep{changagent}. For task planning and execution, RS-Agent converts user intent into executable sequences of remote sensing tool operations~\citep{xu2026rsagent}. Earth-Agent performs quantitative Earth observation analysis through multi-step tool calls and working-memory updates~\citep{feng2026earthagent}. To handle more complex analysis, recent work introduces multi-agent collaboration. GeoMMAgent and GeoLLM-Squad divide knowledge retrieval, target perception, and integrated reasoning among several agents~\citep{xiao2026geomm,lee2025geollmsquad}. GaiaAgent combines a vision-language change detection model with five specialized agents for multi-task land cover change detection and quantitative analysis~\citep{wang2026gaiaagent}. For long-horizon tasks, HiRS-Agent uses a hierarchical architecture with dynamic replanning~\citep{hirsagent2026}. RingMo-Agent and RingMo-Claw also explore autonomous execution and capability extension for remote sensing tasks~\citep{hu2026ringmoagent,kang2026ringmoclaw}.

Tool organization and the reuse of past experience are another research focus. GeoLLM-Engine integrates geospatial APIs, map operators, and external knowledge into a unified tool environment~\citep{singh2024geollmengine}. OpenEarthAgent integrates a remote sensing tool chain and builds standardized tool-call trajectories~\citep{shabbir2026openearthagent}. For tool retrieval and selection, RS-Claw retrieves the required tools level by level through a skill tree~\citep{rsclaw2026}. RemoteAgent decides by itself whether to reason directly or to call specialized tools~\citep{remoteagent2026}. REMSA selects suitable foundation models under given constraints~\citep{chen2025remsa}. OpenEarth-Agent further explores creating tools on demand~\citep{zhao2026openearthagent}. In addition, RSMeM, GeoEvolver, and GeoForge use past trajectories and execution experience to improve subsequent task solving~\citep{wu2026rsmem,dai2026geoevolver,xiaox2026geoforge}.

Agents are also built for specific applications. RescueADI decomposes disaster interpretation into coordinated perception and analysis tasks~\citep{rescueadi2025}. AirSpatialBot targets the recognition and retrieval of fine-grained vehicle attributes in aerial images~\citep{airspatialbot2025}. MineAgent and PEACE are designed for mineral exploration and geological map understanding, respectively~\citep{yu2024mineagent,huang2025peace}. Agents have further been applied to satellite observation scheduling. TSLC uses large language models (LLMs) to coordinate order-driven satellite task planning~\citep{chen2026tslc}. For large-scale agile Earth observation satellites, LLM agents are used to solve bi-level online order dispatching and dynamic scheduling~\citep{zhang2026llmoods}.

Existing benchmarks evaluate remote sensing agents in terms of task completion, tool use, and reasoning. ThinkGeo focuses on image-driven tool chains, multi-step planning, and spatial reasoning~\citep{shabbir2025thinkgeo}. EarthBench covers quantitative Earth observation tasks on multi-source spectral data and remote sensing products~\citep{feng2026earthagent}. TerraLogic examines hierarchical geospatial reasoning and compositional analysis~\citep{yan2026terralogic}. GeoMMBench systematically evaluates perception, knowledge use, and reasoning in multimodal geoscience tasks~\citep{xiao2026geomm}. GeoLLM-Engine, UnivEARTH, and OpenEarth-Bench build diverse geospatial tool environments for evaluation~\citep{singh2024geollmengine,kao2025univearth,zhao2026openearthagent}. TerraBench further studies agent reasoning over heterogeneous Earth system data~\citep{nguyen2026terrabench}. For specialized tasks, AirSpatialBot and PEACE provide benchmarks for aerial spatial question answering and geological map understanding, respectively~\citep{airspatialbot2025,huang2025peace}. CHOICE evaluates perception and reasoning in remote sensing VQA~\citep{li2025choice}.

In summary, prior work on remote sensing agents mainly focuses on task planning, tool selection, knowledge retrieval, and workflow execution. Less attention has been paid to whether the observations returned by normally executed tools are reliable. \citet{munir2026agenticrs} note that remote sensing workflows operate on georeferenced, multimodal, and temporally structured data. Reprojection, resampling, mosaicking, and aggregation change the underlying state and constrain subsequent analysis, so errors can propagate through subsequent reasoning. We quantify the propagation of tool observation errors across ThinkGeo, EarthBench, and TerraLogic, demonstrating their significant impact on remote sensing agents.

\subsection{Agent Verification}
\label{sec:agent_tool_verification}

Verification for general agents covers three levels: model reasoning, tool-call actions, and returned results~\citep{yu2025trustworthyagents,yehudai2026agenteval}. At the reasoning level, Reflexion~\citep{reflexion}, RetroFormer~\citep{retroformer}, and Self-Refine~\citep{madaan2023selfrefine} correct generated content with self-reflection feedback or external evaluators. SCoRe~\citep{kumar2025score} studies optimization with process rewards. Other work analyzes when self-correction works~\citep{huang2024selfcorrect}. CRITIC~\citep{gou2024critic} guides the correction of multimodal reasoning through interactive tool use. SPORT~\citep{li2025sport} assigns a process reward to each reasoning step to guide trajectory optimization. SGV~\citep{andrade2026sgv} addresses bias when multimodal models evaluate agent behavior. It first forms an independent expectation of the behavior and then uses it to evaluate candidate trajectories. RV-LLM~\citep{zhang2025rvllm} proposes a runtime verification framework to detect hallucinations. Tool-MVR~\citep{ma2025toolmvr} trains models on verified tool-call data and reflection samples to improve tool calling and error correction. Other studies analyze fault propagation and failure attribution in long-horizon reasoning~\citep{huang2025faultyagents,zhang2025failureattribution,zhang2026pixelcraft,gao2024clova}.

At the action level, ToolEmu~\citep{toolemu} evaluates the potential risks of tool calls. ToolSword~\citep{ye2024toolsword} detects malicious tool-call intentions. Relign~\citep{xu2025relign} addresses hallucinations in tool selection and use by introducing actions such as deferring execution, asking for clarification, and changing the selected tool. Vigil~\citep{lin2026vigil} and ToolGate~\citep{liu2026toolgate} check the completeness and safety of arguments before a call. AgentSpec~\citep{agentspec}, AgentLTL~\citep{agentltl}, Reason Less, Verify More~\citep{reasonlessverifymore}, Reinforced Agent~\citep{reinforcedagent}, and Verified Tool Calls~\citep{verifiedtoolcalls} constrain execution with temporal logic rules or formal specifications. At the result level, Tools Fail~\citep{sun2024toolsfail} studies how to detect silent tool errors. RIVA~\citep{abuzakuk2026riva} cross-checks the results of different tools to reduce the effect of wrong outputs on detection. EG-VAR~\citep{egvar} requires reasoning to rest on traceable evidence. Trust or Escalate~\citep{trustorescalate} specifically enables agents to abstain or escalate when the available evidence is determined to be insufficient.

Existing remote sensing agents mainly improve execution reliability through feedback and rule checks. GeoMMAgent decides whether to output a conclusion or re-execute based on self-evaluation~\citep{xiao2026geomm}. HiRS-Agent introduces step-level verification and replanning~\citep{hirsagent2026}. CangLing-KnowFlow constrains workflows with tool preconditions and quality feedback, and triggers repair when anomalies occur~\citep{chen2025canglingknowflow}. In satellite task scheduling, TSLC checks feasibility using external rules~\citep{chen2026tslc}.

In summary, existing studies verify agents at the level of reasoning, actions, and execution results. Unlike work that verifies agent execution steps, checks tool outputs against images and task context, or cross-checks results between tools~\citep{hirsagent2026,sun2024toolsfail,abuzakuk2026riva}, we focus on the reliability of the observations returned by remote sensing tools. Remote sensing agents rely on visual perception, spatial measurement, and raster analysis tools, and most of their outputs are uncertain observations. Even under normal execution, these observations can be wrong because of scene complexity, image quality, and model prediction uncertainty. The errors can then propagate through subsequent reasoning. \ours{} builds its verification on two intuitions. First, exposing the internal process of a tool helps judge whether an observation is trustworthy. We therefore define a verifiable observation protocol that requires tools to provide process evidence. Second, the past performance of different tool configurations on specific tasks can serve as a reference for verification. We therefore build a task--tool reliability prior. The framework verifies tool observations with both the process evidence and the prior, limiting the propagation of incorrect observations.

\section{Empirical Analysis of Tool Observations}
\label{sec:empirical_analysis}

To quantify the role of tool observations in the reasoning of remote sensing agents, we collect 1,229 execution trajectories from the reference baseline agents of EarthBench, ThinkGeo, and TerraLogic, namely Earth-Agent~\citep{feng2026earthagent}, ThinkGeo-Agent~\citep{shabbir2025thinkgeo}, and HieraPlan~\citep{yan2026terralogic}. We study three questions. First, how often do agents depend on tool observations? Second, how often are tool observations wrong when the tool executes normally? Third, how often do incorrect observations propagate to the final answer? In this offline analysis, observation errors are judged against the expert reference trajectories and ground-truth answers from the benchmarks. These references are used only for evaluation and are never given to the online verifier.

\subsection{Metric Definitions}
\label{sec:empirical_definition}

For a task set $T$, we define three task-level metrics.

\textbf{Observation dependence rate} $\mathcal{R}_{\mathrm{dep}}$ measures the fraction of tasks that use tool observations. Let $T_{\mathrm{dep}}\subseteq T$ be the subset of tasks in which at least one tool observation is used in subsequent reasoning. Then
{\small
\begin{equation}\mathcal{R}_{\mathrm{dep}}=\frac{|T_{\mathrm{dep}}|}{|T|}.
\end{equation}
}

\textbf{Observation error rate} $\mathcal{R}_{\mathrm{err}}$ measures how often observations are wrong when tools execute normally. An observation error means that the tool completes execution and reports no error, but the returned observation disagrees with the reference facts. Let $T_{\mathrm{err}}\subseteq T_{\mathrm{dep}}$ be the subset of tasks with at least one observation error. Then
{\small
\begin{equation}\mathcal{R}_{\mathrm{err}}=\frac{|T_{\mathrm{err}}|}{|T_{\mathrm{dep}}|}.
\end{equation}
}
This metric separates the execution status of a tool from the actual correctness of its observation.

\textbf{Error propagation rate} $\mathcal{R}_{\mathrm{prop}}$ measures how often a wrong observation is adopted by subsequent reasoning and carried into a wrong final answer. Let $T_{\mathrm{prop}}\subseteq T_{\mathrm{err}}$ denote the tasks where a wrong observation is adopted by subsequent reasoning and the wrong final answer is consistent with that observation. Then
{\small
\begin{equation}\mathcal{R}_{\mathrm{prop}}=\frac{|T_{\mathrm{prop}}|}{|T_{\mathrm{err}}|}.
\end{equation}
}
This definition corresponds to the error propagation probability that AgentProp-Bench~\citep{agentpropbench} measures for general tool agents. The difference is that we count naturally occurring errors rather than injected ones.

\subsection{Results}
\label{sec:empirical_results}

\begin{table}[t]
\centering
\caption{Tool observation statistics (\%) of the benchmark reference agents.}
\label{tab:empirical_stats}
\footnotesize
\setlength{\tabcolsep}{4pt}
\begin{tabular}{lllccc}
\toprule
\textbf{Benchmark} & \textbf{Agent} & \textbf{Backbone} & $\mathcal{R}_{\mathrm{dep}}$ & $\mathcal{R}_{\mathrm{err}}$ & $\mathcal{R}_{\mathrm{prop}}$ \\
\midrule
EarthBench & Earth-Agent & GPT-5 & 88.7 & 22.7 & 82.0 \\
ThinkGeo   & ThinkGeo-Agent    & GPT-4o & 91.7 & 41.3 & 91.5 \\
TerraLogic & HieraPlan   & GPT-4o & 88.1 & 39.8 & 93.2 \\
\bottomrule
\end{tabular}
\end{table}

Table~\ref{tab:empirical_stats} reports the statistics. We report three findings.

\textbf{Remote sensing agents widely depend on tool observations.} $\mathcal{R}_{\mathrm{dep}}$ is at least 88.1\% on all three benchmarks. Tool observations are used in most tasks, indicating that remote sensing agents widely depend on them.

\begin{figure}[!t]
\centering
\includegraphics[width=\columnwidth]{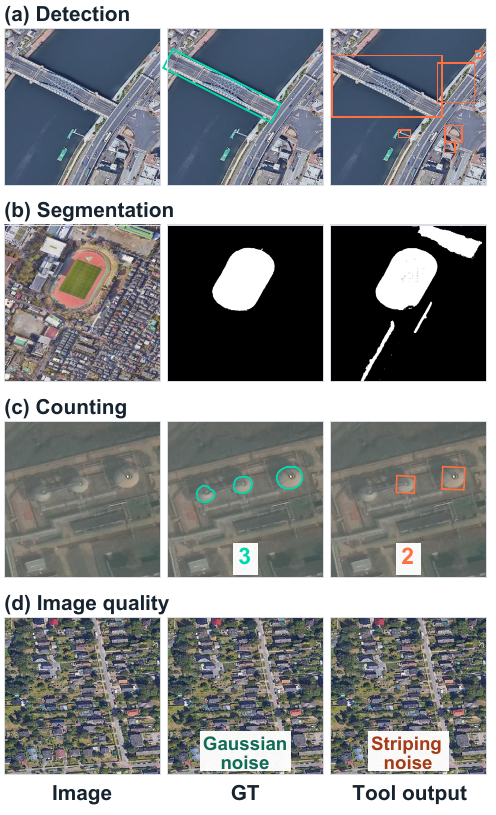}
\caption{Examples of incorrect tool observations under normal execution across (a) detection, (b) segmentation, (c) counting, and (d) image quality.}
\label{fig:observation_case}
\end{figure}

\textbf{Normal execution does not guarantee correct observations.} $\mathcal{R}_{\mathrm{err}}$ is 22.7\%, 41.3\%, and 39.8\% on EarthBench, ThinkGeo, and TerraLogic, respectively. All these errors occur when the tool reports no anomaly, as illustrated in Fig.~\ref{fig:observation_case}. Error handling based on execution status cannot catch them.

\textbf{Wrong tool observations propagate at a high rate.} Among tasks with observation errors, $\mathcal{R}_{\mathrm{prop}}$ is at least 82.0\%, and it exceeds 91\% on ThinkGeo and TerraLogic. Incorrect observations are often carried forward by subsequent reasoning. This is consistent with over-trust in general tool agents~\citep{sun2024toolsfail,whenthetooldecides,toolgatevlm}.

\begin{figure*}[!t]
\centering
\includegraphics[width=\textwidth]{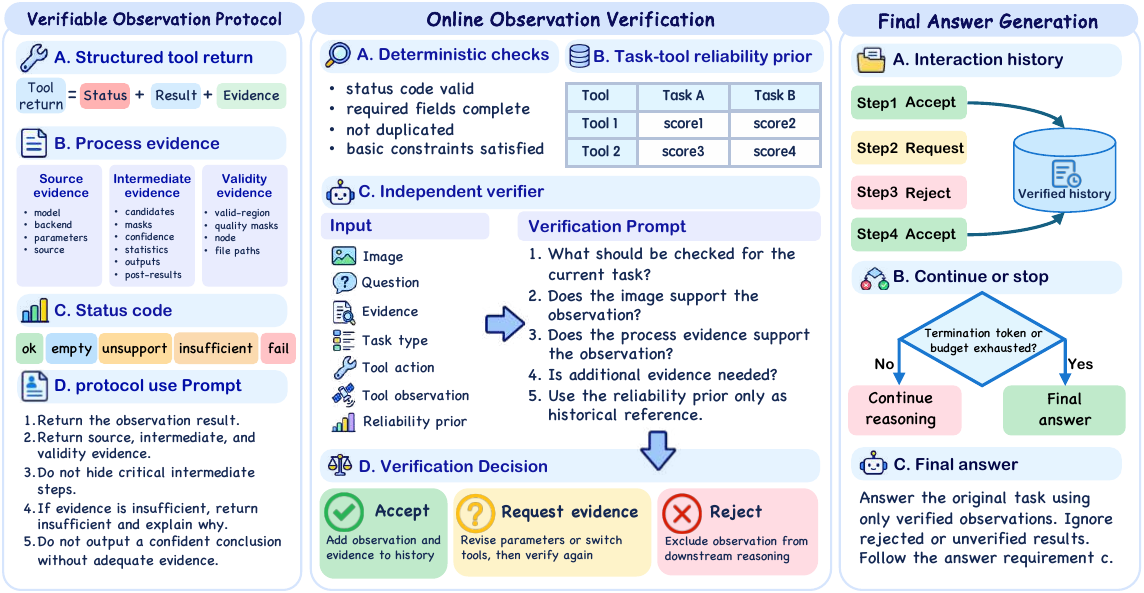}
\caption{Overall framework of \ours{}.
The verifiable observation protocol requires tools to return execution status, observations, and process evidence (source evidence, intermediate results, and validity evidence). Online verification applies deterministic rule checks, then verifies observations that pass using the image, task, process evidence, and task--tool reliability prior. The verifier decides to accept, request evidence, or reject. Accepted observations and evidence enter the verified history. The backbone model uses feedback to plan subsequent tool calls and generates the final answer from this history after interaction ends.}
\label{fig:framework}
\end{figure*}

\section{Method}
\label{sec:method}

To address the propagation of tool observation errors identified in Section~\ref{sec:empirical_analysis}, we propose \ours{}, an observation verification framework. It verifies each tool observation after the tool executes and before the backbone model reads the observation. Fig.~\ref{fig:framework} shows the overall architecture. The verifiable observation protocol requires tools to return the execution status and process evidence with each observation. Online verification first runs deterministic rule checks and then queries the reliability prior for the task and tool configuration. The verifier uses the same model as the backbone but a separate context. It combines the input images, the process evidence, and the prior to accept the observation, request additional evidence, or reject it. Accepted observations enter the verified history for final reasoning, while ongoing tasks continue planning and tool execution based on feedback.

\subsection{Problem Formulation}
\label{sec:base}

A remote sensing task is denoted by $x=(\mathcal{I},q,c)$, where $\mathcal{I}$ is the set of input remote sensing images, $q$ is the natural-language question, and $c$ is the answer requirement that specifies the output format. The agent follows the ReAct framework, in which a backbone model $M$ selects tools and generates call arguments. The initial history $h_0$ contains the task input $x$, the system prompt, and the tool instructions.

The tool set $\mathcal{K}$ contains four types of remote sensing tools.
\texttt{perceive} provides 7 perception modes, such as detection and segmentation.
\texttt{measure} provides 7 metrics, such as count and area.
\texttt{execute} runs typed DAGs and includes 10 built-in standard Earth observation products. Its functions cover product computation, raster algebra, statistical analysis, time series, and spatial relation analysis. \texttt{retrieve} provides 3 types of knowledge retrieval for algorithm specifications and domain knowledge. The first three tools return perception and quantitative results about ground objects, which we define as tool observations $o_t$. \texttt{retrieve} returns auxiliary text knowledge cards. \ref{app:tools} provides tool descriptions and arguments; \ref{app:cards} provides the retrieval card library.

The total tool-call budget is $H_{\max}=10$, including first calls and retries. Before the $t$-th tool call, the backbone model reads the history $h_{t-1}$ and generates an action $a_t=(k_t,u_t)$, where $k_t\in\mathcal{K}$ is the selected tool and $u_t$ is the call argument. The environment executes the action and returns a result $R_t$. The base ReAct framework updates the history directly:
{\small
\begin{equation}\label{eq:base_history}
h_t=h_{t-1}\oplus(a_t,R_t),
\end{equation}
}
where $\oplus$ appends action and result to the history context.

The interaction ends when the backbone model outputs a stop token or the call budget runs out. The actual number of tool calls is $H\le H_{\max}$. In the base ReAct framework, wrong tool observations also enter the reasoning context and may affect subsequent reasoning. \ours{} verifies each tool observation before the history update and decides whether to add it to the reasoning context.

\subsection{Verifiable Observation Protocol}
\label{sec:vop}

Verifying a tool observation requires checking its data source, computation process, and applicable conditions. \ours{} therefore defines a unified verifiable observation protocol for the three observation tools. Besides the observation, each call returns the execution status and process evidence:
{\small
\begin{equation}\label{eq:return}
R_t=(\sigma_t,o_t,\mathcal{E}_t),\qquad
\mathcal{E}_t=(\mathcal{E}^{\mathrm{src}}_t,
\mathcal{E}^{\mathrm{mid}}_t,
\mathcal{E}^{\mathrm{val}}_t),
\end{equation}
}
where $\sigma_t\in\{\text{ok},\text{empty},\text{unsupport},
\text{insufficient},\text{fail}\}$ is the execution status code,
$o_t$ is the tool observation, and $\mathcal{E}_t$ is the process evidence.

Process evidence has three parts, as summarized in Table~\ref{tab:tools}. Source evidence $\mathcal{E}^{\mathrm{src}}_t$ records the backend model, thresholds, sensor type, and input resolution. Intermediate results $\mathcal{E}^{\mathrm{mid}}_t$ record target polygons, confidence scores, the numerator and denominator of a measurement, and the number of valid pixels. Validity evidence $\mathcal{E}^{\mathrm{val}}_t$ records valid-pixel masks, class coverage flags, and paths to intermediate files.

\begin{table*}[t]
\centering
\small
\caption{Process evidence returned by the three observation tools.}
\label{tab:tools}

\setlength{\tabcolsep}{6pt}
\renewcommand{\arraystretch}{1.12}
\newcommand{\toolcolwidth}{%
    \dimexpr(\textwidth-3.6cm-8\tabcolsep)/3\relax
}

\begin{tabular}{
    w{l}{3.6cm}
    w{c}{\toolcolwidth}
    w{c}{\toolcolwidth}
    w{c}{\toolcolwidth}
}
\toprule
Evidence type &
\texttt{perceive} &
\texttt{measure} &
\texttt{execute} \\
\midrule

\makecell[cl]{
    Source evidence $\mathcal{E}^{\mathrm{src}}_t$
} &
\makecell[cc]{
    Detector weights \\
    Thresholds \\
    Sensor \\
    Resolution
} &
\makecell[cc]{
    Input layer sources \\
    Metric \\
    Unit specification
} &
\makecell[cc]{
    Operator parameters \\
    Product formulas \\
    Band mapping
} \\
\midrule

\makecell[cl]{
    Intermediate results $\mathcal{E}^{\mathrm{mid}}_t$
} &
\makecell[cc]{
    Object bounding boxes \\
    Confidence scores \\
    Number of filtered candidates
} &
\makecell[cc]{
    Measurement numerator \\
    Denominator \\
    Pixel counts
} &
\makecell[cc]{
    Per-node outputs \\
    Valid-pixel counts
} \\
\midrule

\makecell[cl]{
    Validity evidence $\mathcal{E}^{\mathrm{val}}_t$
} &
\makecell[cc]{
    Layer status \\
    Class completeness \\
    Geometry flags
} &
\makecell[cc]{
    Layer status \\
    Class coverage
} &
\makecell[cc]{
    Node status \\
    Quality masks \\
    List of failed nodes
} \\
\bottomrule
\end{tabular}
\end{table*}

The protocol requires tools to return all required evidence. Even when the status is \texttt{ok}, an observation with missing required fields cannot pass verification. When data quality is insufficient, the tool returns \texttt{insufficient} with a reason and does not output an estimate.

Tools save derived layers and intermediate data as local files so that programs can re-check them. Coordinates predicted by vision-language models without spatial calibration are marked as approximate geometry. Such coordinates cannot be used by \texttt{measure} for quantitative geometric measurement. Field descriptions are given in \ref{app:tools}, and the prompts for each perception mode are given in \ref{app:vlm_prompts}.

\subsection{Task--Tool Reliability Prior}
\label{sec:prior}

The deterministic rules used here check execution status and computation conditions, but do not directly assess whether an observation matches the image content. Tool performance also varies with the task type. For example, an open-vocabulary perception tool can judge whether a target exists but may miss objects when counting dense small targets. Because online verification cannot access the current observation's ground truth, \ours{} uses offline calibration results as a historical reference. The prior reflects how a tool configuration has performed on the task type and helps the verifier check current evidence.

We define the task-type set $\mathcal{Z}$ using the 21 multiple-choice task types of CHOICE~\citep{li2025choice}, so $|\mathcal{Z}|=21$. The set includes image captioning and excludes visual grounding and referring expression segmentation. Definitions of all task types are given in \ref{app:choice_tasks}. The calibration set $\mathcal{D}_{\mathrm{cal}}$ contains 1,200 tasks from OpenEarthAgent~\citep{shabbir2026openearthagent} and CHOICE. The evaluation set CHOICE-420 is a subset of CHOICE. The calibration set does not overlap with the test set of any evaluation benchmark.

The original task labels differ across data sources. The system maps each task to a type $z\in\mathcal{Z}$ based on its question $q$. It first applies deterministic keyword rules and falls back to a large language model when no rule matches. Offline calibration and online verification use the same classification procedure.

With GPT-4o as the backbone, the system runs the base ReAct framework on the calibration set. Each calibration task is run once with at most $H_{\max}=10$ tool calls. This yields 10,795 tool-call records for estimating the past performance of each configuration.
Counting calls only by tool type ignores differences in function and configuration. We therefore count calls at both a coarse and a fine level. Coarse configurations distinguish perception modes, measurement metrics, and execution operators. Fine perception configurations further distinguish the backend and the view setting, and fine measurement configurations further distinguish the spatial relation and the input layer type. Each level aggregates its own call counts and computes priors via Eq.~\eqref{eq:reliability_prior}. Configuration examples and statistics are given in \ref{app:reliability_prior}.

For task type $z$ and configuration $v$, $n(z,v)$ is the total number of past calls. $n_{\mathrm{valid}}(z,v)$ is the number of these calls that executed normally, had a valid format, and belonged to a task whose final answer was correct. We use task-level correctness as an offline weak supervision signal. All calls in a task share the outcome of that task. The resulting score therefore reflects how a tool configuration has performed on task type $z$. We estimate reliability with Laplace smoothing:
{\small
\begin{equation}\label{eq:reliability_prior}
\hat{r}(z,v)=\frac{n_{\mathrm{valid}}(z,v)+1}{n(z,v)+2}.
\end{equation}
}
This estimate avoids scores of exactly 0 or 1 when there are few samples.

The scores of all task types and tool configurations form a task--tool reliability matrix. The matrix is fixed after offline calibration and is only queried during online execution. Table~\ref{tab:matrix_rows} lists selected entries. More statistics and configuration identifiers are given in \ref{app:reliability_prior}.

\begin{table}[t]
\centering
\footnotesize
\caption{Selected entries of the task--tool reliability prior.}
\label{tab:matrix_rows}

\setlength{\tabcolsep}{2pt}
\renewcommand{\arraystretch}{1.08}
\renewcommand{\tabularxcolumn}[1]{p{#1}}

\begin{tabularx}{\columnwidth}{
    l
    l
    >{\raggedright\arraybackslash}X
    c
    c
}
\toprule
Task type & Tool & Configuration & Level & $\hat{r}$ \\
\midrule
Object Counting &
\texttt{perceive} &
\texttt{objects} &
Coarse & 0.926 \\

Object Counting &
\texttt{measure} &
\texttt{count|auto|objects} &
Fine & 0.975 \\

Object Presence &
\texttt{perceive} &
\texttt{objects} &
Coarse & 0.800 \\

Image Caption &
\texttt{perceive} &
\texttt{evidence} &
Coarse & 0.923 \\

Change Detection &
\texttt{perceive} &
\texttt{change} &
Coarse & 0.321 \\

Env. Assessment &
\texttt{execute} &
\texttt{raster.product} &
Coarse & 0.810 \\

Env. Assessment &
\texttt{execute} &
\texttt{asset.catalog} &
Coarse & 0.803 \\

Physical Property &
\texttt{measure} &
\texttt{area|auto|regions} &
Fine & 0.706 \\
\bottomrule
\end{tabularx}
\end{table}

For the $t$-th tool call, the system determines the fine configuration $v_t^{\mathrm{fine}}$ and the coarse configuration $v_t^{\mathrm{coarse}}$ from the action $a_t$. The prior retrieved for this call is denoted by $\hat{r}_t$. During offline construction, the matrix keeps only entries with at least $n_{\min}=8$ samples. At query time, the system first looks up the fine entry and falls back to the coarse entry if no fine entry is available. If neither level has an entry, the system returns \texttt{None}. The lookup rule is
{\small
\begin{equation}\label{eq:hierarchical_prior}
\hat{r}_t=
\begin{cases}
\hat{r}(z,v_t^{\mathrm{fine}}),
& n(z,v_t^{\mathrm{fine}})\ge n_{\min},\\
\hat{r}(z,v_t^{\mathrm{coarse}}),
& n(z,v_t^{\mathrm{fine}})<n_{\min}
\le n(z,v_t^{\mathrm{coarse}}),\\
\text{None},
& \text{otherwise}.
\end{cases}
\end{equation}
}

Uncalibrated configurations still go through full verification, but semantic verification uses no prior information for them. All backbone models in our evaluation share the same offline reliability matrix. Section~\ref{sec:ablation} uses GPT-5 as the backbone and examines the added value of the prior through ablation.

\subsection{Online Verification}
\label{sec:verify}

Online verification applies to the observations returned by \texttt{perceive}, \texttt{measure}, and \texttt{execute}. The system first runs deterministic rule checks and then performs semantic verification on the observations that pass. The verification result is to accept the observation, request additional evidence, or reject it, denoted by \texttt{accept}, \texttt{request\_evidence}, and \texttt{reject}.

For the $t$-th call that passes the rule checks, the semantic verification result is $V_t$:
{\small
\begin{equation}\label{eq:verify_joint}
V_t\in
\{\texttt{accept},\texttt{request\_evidence},\texttt{reject}\}.
\end{equation}
}

The deterministic rule checks are run by programs without calling a language model:
{\small
\begin{equation}\label{eq:det_check}
V_t^{\mathrm{det}}=
\mathbb{I}\Bigl(
(\sigma_t=\texttt{ok})
\wedge\operatorname{Valid}(a_t,\mathcal{E}_t)
\wedge\operatorname{Novel}(a_t,h_{t-1})
\Bigr),
\end{equation}
}
where $\mathbb{I}(\cdot)$ is the indicator function. $\operatorname{Valid}$ checks required fields, geometric coordinates, and numerical statistics according to the tool type and mode. For computations that require a non-empty region, it also checks that the mask area is positive. $\operatorname{Novel}$ compares the arguments and input image extent of the current call with those of past calls to rule out repeated calls.

If any check fails, the system returns a diagnosis $f_t^{\mathrm{det}}$ so that the backbone model can adjust the call. The current observation does not enter subsequent reasoning.

After the rule checks pass, the verifier $M_{\mathrm{eval}}$ judges whether the image and process evidence support the current observation. $M_{\mathrm{eval}}$ uses the same model as the backbone $M$ but does not share its dialogue context:
{\small
\begin{equation}\label{eq:sem_check}
(V_t,f_t^{\mathrm{sem}})
=
M_{\mathrm{eval}}(\mathcal{I},q,z,a_t,o_t,\mathcal{E}_t,\hat{r}_t),
\end{equation}
}
where $z$ is the task type and $f_t^{\mathrm{sem}}$ is the verification feedback. The verifier infers what to check and what the task requires from the question $q$ and the action $a_t$. For example, for object detection it checks the instance class in each box, and for measurement it checks whether the object matches the question.

Semantic verification uses a separate context and receives only the inputs in Eq.~\eqref{eq:sem_check}. The verifier does not read the interaction history or the earlier reasoning text of the backbone model. During verification, the history is used only for the repeated-call check in the deterministic rules. The verification prompt is given in \ref{app:verify_prompt}.

The reliability prior points the verifier to the past performance of the tool on the current task. When the score is low, the verifier must check more strictly whether the evidence is sufficient. When the score is high, the observation must still be supported by the image and the process evidence. If $\hat{r}_t=\text{None}$, the verifier decides from the other inputs only.

The verifier checks whether the current evidence supports the observation. If the evidence is sufficient, it accepts the observation. If the evidence is insufficient, it requests additional evidence and states what is needed. If the evidence shows that the observation is wrong or inapplicable, it rejects the observation and gives the reason.

For the three observation tools, history updating follows the check and verification results. On acceptance, it appends the current action, the observation, and the process evidence. On an evidence request or a rejection, it appends the action, the verification decision, and the feedback. If the rule checks fail, it appends only the action and the rule diagnosis:
{\small
\begin{equation}\label{eq:state_update}
h_t=
\begin{cases}
h_{t-1}\oplus(a_t,o_t,\mathcal{E}_t),
& V_t^{\mathrm{det}}=1,\ V_t=\texttt{accept},\\
h_{t-1}\oplus(a_t,V_t,f_t^{\mathrm{sem}}),
& V_t^{\mathrm{det}}=1,\ V_t\ne\texttt{accept},\\
h_{t-1}\oplus(a_t,f_t^{\mathrm{det}}),
& V_t^{\mathrm{det}}=0.
\end{cases}
\end{equation}
}

When the verifier requests additional evidence or rejects an observation, the backbone model adjusts the arguments or switches tools based on the feedback $f_t^{\mathrm{sem}}$. Specifically, an evidence request allows at most $K_{\max}=3$ targeted follow-up calls within the remaining budget, whereas a rejection discards the faulty observation and prompts the agent to replan. Each follow-up call counts as a new tool call, for which the prior is queried and verification is run again. The evidence request ends when the observation is accepted or rejected, or when the follow-up limit is reached. If the call budget is not exhausted, the backbone model continues reasoning from the updated history.

We denote the complete verified history used for the final answer by $\mathcal{H}_H^{\mathrm{ver}}$. Its tool observation records retain only the triples $(a_t,o_t,\mathcal{E}_t)$ with $V_t=\texttt{accept}$ (along with any auxiliary knowledge cards from \texttt{retrieve}), in call order.

After the interaction ends, the backbone model generates the final answer in a newly built context. This stage does not reuse the dialogue context from execution. Its input contains the original task description, the answer requirement, and the complete verified history $\mathcal{H}_H^{\mathrm{ver}}$:
{\small
\begin{equation}\label{eq:final_answer}
y=M(x,\mathcal{H}_H^{\mathrm{ver}}).
\end{equation}
}
The final-answer prompt is in \ref{app:final_prompt}. Benchmark input and output requirements are in \ref{app:task_templates}.

\section{Experiments}
\label{sec:experiments}

\subsection{Experimental Setup}
\label{sec:setup}

\begin{figure}[!t]
\centering
\includegraphics[width=\columnwidth]{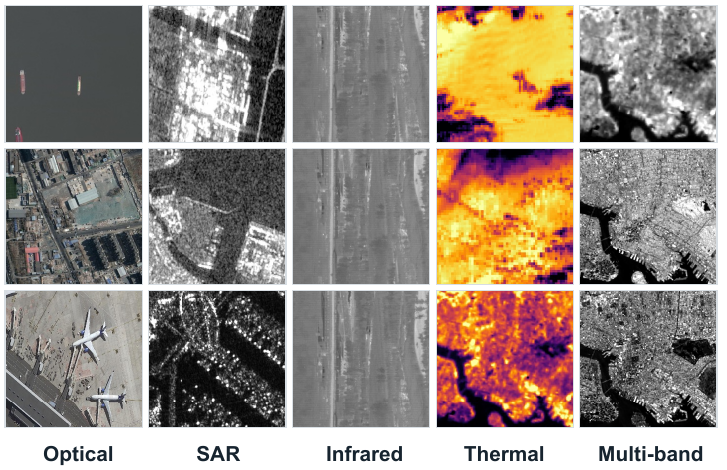}
\caption{Representative remote sensing inputs across the benchmarks, including optical, SAR, infrared, thermal, and multi-band imagery.}
\label{fig:input_case}
\end{figure}

\noindent\textbf{Datasets.}
We evaluate on ThinkGeo~\citep{shabbir2025thinkgeo}, EarthBench~\citep{feng2026earthagent}, and TerraLogic~\citep{yan2026terralogic}, with 436, 248, and 545 test tasks, respectively. ThinkGeo tests structured tool use, multi-step planning, and spatial reasoning on remote sensing images. EarthBench tests autonomous planning and multi-step tool calls for quantitative analysis of RGB images, spectral data, and remote sensing products. TerraLogic focuses on hierarchical geospatial reasoning through long compositional analysis chains. We also construct CHOICE-420 from the 21 multiple-choice task types in CHOICE~\citep{li2025choice}. We exclude the other two of its 23 task types, visual grounding and referring expression segmentation, which require bounding boxes or segmentation masks. This keeps the answer format consistent across the selected tasks. We select 20 samples per type, yielding 280 perception tasks across 14 types and 140 reasoning tasks across 7 types. The resulting benchmark contains 420 tasks, with task definitions provided in \ref{app:choice_tasks}. Fig.~\ref{fig:input_case} shows representative inputs across these benchmarks. The reliability calibration set does not overlap with any evaluation set.

\begin{table*}[t]
\centering
\caption{Task performance (\%) on RS agent benchmarks. $^{\dagger}$ denotes results following RS-Claw's test split on EarthBench.}
\label{tab:main_results}
\small
\setlength{\tabcolsep}{7pt}
\renewcommand{\arraystretch}{1.08}
\begin{tabular}{llccccc}
\toprule
\multirow{2}{*}{Method} &
\multirow{2}{*}{Backbone} &
\multirow{2}{*}{Weights} &
\multicolumn{2}{c}{ThinkGeo} &
EarthBench & TerraLogic \\
\cmidrule(lr){4-5}\cmidrule(lr){6-6}\cmidrule(lr){7-7}
& & & Ans. & $\mathrm{Ans}_I$ & Acc. & AnsAcc \\
\midrule
ThinkGeo-Agent & GPT-4o & Closed & 9.8 & 20.4 & -- & -- \\
HieraPlan & GPT-4o & Closed & -- & -- & -- & 25.7 \\
Earth-Agent & GPT-4o & Closed & -- & -- & 45.3 & -- \\
Earth-Agent & GPT-5 & Closed & -- & -- & 65.6 & -- \\
OpenEarth-Agent & GPT-5 & Closed & -- & -- & 67.6 & -- \\
RS-Claw$^{\dagger}$ & GPT-5 & Closed & -- & -- & 68.7 & -- \\
HiRS-Agent & Qwen3-4B & Open & 11.3 & 13.8 & 44.0 & -- \\
HiRS-Agent & Qwen3-8B & Open & 12.1 & 14.8 & 48.4 & -- \\
\midrule
\multirow{13}{*}{\ours{}}
& Qwen3.5-27B & Open & 16.9 & 24.1 & 59.3 & 27.7 \\
& Qwen3.5-122B & Open & 18.7 & 25.3 & 60.5 & 29.0 \\
& Qwen3.5-397B & Open & 22.1 & 30.0 & 65.7 & 29.7 \\
& Mistral Small 4 & Open & 13.5 & 21.7 & 54.8 & 25.9 \\
& Qwen3.6-27B & Open & 20.6 & 27.4 & 65.3 & 28.3 \\
& GLM-4.6V & Open & 22.5 & 29.8 & 66.1 & 30.1 \\
& Kimi K2.6 & Open & 23.2 & 31.6 & 66.9 & 31.6 \\
\cmidrule(lr){2-7}
& Gemini-3.5-Flash & Closed & 21.9 & 30.6 & 64.9 & 30.5 \\
& Gemini-3.1-Pro & Closed & 23.9 & 32.8 & 66.5 & 32.1 \\
& GPT-4o & Closed & 14.1 & 22.6 & 55.2 & 26.1 \\
& GPT-5 & Closed & 23.6 & 32.6 & 68.1 & 31.4 \\
& GPT-5$^{\dagger}$ & Closed & 23.6 & 32.6 & 69.2 & 31.4 \\
& GPT-5.5 & Closed &
\textbf{26.4} & \textbf{32.9} & \textbf{70.2} & \textbf{33.0} \\
\bottomrule
\end{tabular}
\end{table*}

\noindent\textbf{Evaluation Protocols and Compared Methods.}
For ThinkGeo, we follow the official protocol, using GPT-4o-mini to judge answer correctness based on each query's evaluation question. We report final answer accuracy (Ans.) and image-grounded answer accuracy ($\mathrm{Ans}_I$), both of which are end-to-end metrics. For EarthBench, we report end-to-end accuracy (Acc.) in Autonomous Planning (AP) mode. For TerraLogic, we report final answer accuracy (AnsAcc). On CHOICE-420, we report Overall, Perception, and Reasoning accuracy. The compared methods include ThinkGeo-Agent~\citep{shabbir2025thinkgeo}, HieraPlan~\citep{yan2026terralogic}, Earth-Agent~\citep{feng2026earthagent}, OpenEarth-Agent~\citep{zhao2026openearthagent}, HiRS-Agent~\citep{hirsagent2026}, and RS-Claw~\citep{rsclaw2026}. We use results from their original papers, including both accuracy and efficiency for Earth-Agent. On CHOICE-420, direct answering by each backbone model serves as the baseline for measuring accuracy changes with the full framework.

\noindent\textbf{Implementation Details.}
We obtain the base ReAct configuration \base{} by disabling observation verification and its history management and removing the task--tool reliability prior. \base{} passes tool results directly to subsequent reasoning and shares the backbone model, tool set, and call budget with \ours{}. The full framework combines the verifiable observation protocol with online verification and history management, querying the prior during verification. Following Section~\ref{sec:method}, we set the tool-call limit to $H_{\max}=10$ and allow at most $K_{\max}=3$ follow-up calls per observation. All follow-up calls count toward the total budget. The verifier uses the same model as the backbone but operates in a separate context. We build the reliability prior from the offline calibration set, requiring at least $n_{\min}=8$ past samples, and keep it fixed during evaluation. Tool arguments and prompts are provided in \ref{app:tools} and \ref{app:prompts}, respectively.

\begin{table*}[t]
\centering
\caption{Accuracy (\%) on CHOICE-420: direct answering (Origin) versus \ours{}. Parentheses show gains in percentage points. Bold marks the best value.}
\label{tab:choice420}
\small
\setlength{\tabcolsep}{7pt}
\renewcommand{\arraystretch}{1.08}
\begin{tabular}{lccccccc}
\toprule
\multirow{2}{*}{Backbone} &
\multirow{2}{*}{Weights} &
\multicolumn{2}{c}{Overall} &
\multicolumn{2}{c}{Perception} &
\multicolumn{2}{c}{Reasoning} \\
\cmidrule(lr){3-4}\cmidrule(lr){5-6}\cmidrule(lr){7-8}
& & Origin & \ours{} & Origin & \ours{} & Origin & \ours{} \\
\midrule
Mistral Small 4 & \multirow{7}{*}{Open}
& 75.71 & 84.05 (+8.34)
& 76.07 & 86.07 (+10.00)
& 75.00 & 80.00 (+5.00) \\
Qwen3.5-27B &
& 74.05 & 83.33 (+9.28)
& 78.57 & 86.79 (+8.22)
& 65.00 & 76.43 (+11.43) \\
Qwen3.5-122B &
& 69.29 & 82.86 (+13.57)
& 72.50 & 86.79 (+14.29)
& 62.86 & 75.00 (+12.14) \\
Qwen3.5-397B &
& 83.81 & 85.95 (+2.14)
& 85.36 & 87.50 (+2.14)
& 80.71 & 82.86 (+2.15) \\
Qwen3.6-27B &
& 79.76 & 85.24 (+5.48)
& 83.93 & 88.21 (+4.28)
& 71.43 & 79.29 (+7.86) \\
GLM-4.6V &
& 79.29 & 85.71 (+6.42)
& 80.36 & 87.50 (+7.14)
& 77.14 & 82.14 (+5.00) \\
Kimi K2.6 &
& 84.29 & 86.43 (+2.14)
& 87.50 & 89.64 (+2.14)
& 77.86 & 80.00 (+2.14) \\
\midrule
Gemini-3.5-Flash & \multirow{4}{*}{Closed}
& 84.52 & 87.14 (+2.62)
& 88.57 & 90.00 (+1.43)
& 76.43 & 81.43 (+5.00) \\
Gemini-3.1-Pro &
& \textbf{87.14} & \textbf{89.05} (+1.91)
& \textbf{89.64} & 91.07 (+1.43)
& 82.14 & 85.00 (+2.86) \\
GPT-5 &
& 80.48 & 88.10 (+7.62)
& 81.07 & \textbf{91.79} (+10.72)
& 79.29 & 80.71 (+1.42) \\
GPT-5.5 &
& 85.24 & 88.57 (+3.33)
& 86.43 & 88.93 (+2.50)
& \textbf{82.86} & \textbf{87.86} (+5.00) \\
\bottomrule
\end{tabular}
\end{table*}

\subsection{Main Results}
\label{sec:results}

Table~\ref{tab:main_results} reports the results on the three agent benchmarks. With GPT-4o, \ours{} achieves 14.1\% Ans.\ and 22.6\% $\mathrm{Ans}_I$ on ThinkGeo, which are 4.3 and 2.2 percentage points higher than ThinkGeo-Agent. On EarthBench, \ours{} reaches 55.2\%, 9.9 percentage points higher than Earth-Agent with the same backbone. On TerraLogic, \ours{} reaches 26.1\%, 0.4 percentage points higher than HieraPlan.

HieraPlan reports results only with GPT-4o. With GPT-5 and GPT-5.5, \ours{} reaches 31.4\% and 33.0\% AnsAcc on TerraLogic, respectively. With GPT-5, \ours{} reaches 68.1\% accuracy on the full EarthBench test set. This is 2.5 percentage points higher than Earth-Agent and higher than OpenEarth-Agent with Full Tools. Following RS-Claw's test split, \ours{} with GPT-5 reaches 69.2\%, exceeding RS-Claw with the same backbone by 0.5 percentage points. With GPT-5.5, \ours{} reaches 70.2\% on full EarthBench.

Across all \ours{} configurations, $\mathrm{Ans}_I$ is 6.5 to 9.0 percentage points higher than Ans. Under the same framework, performance varies across backbone models. For Qwen3.5, all four metrics increase as the model grows from 27B to 397B. Ans.\ rises from 16.9\% to 22.1\%, and EarthBench accuracy rises from 59.3\% to 65.7\%. With GPT-5.5, the four metrics reach 26.4\%, 32.9\%, 70.2\%, and 33.0\%. These are the highest full-set values in Table~\ref{tab:main_results}. The open-weight backbone Kimi K2.6 reaches 23.2\%, 31.6\%, 66.9\%, and 31.6\%, close to Gemini-3.1-Pro's 23.9\%, 32.8\%, 66.5\%, and 32.1\%. Their relative performance varies across benchmarks. Kimi K2.6 slightly outperforms Gemini-3.1-Pro on EarthBench, while Gemini-3.1-Pro performs better on ThinkGeo and TerraLogic.

\begin{table}[t]
\centering
\caption{Ablation study (\%, GPT-5 backbone).}
\label{tab:ablation}
\footnotesize
\setlength{\tabcolsep}{4.5pt}
\renewcommand{\arraystretch}{1.1}
\begin{tabular}{cccccc}
\toprule
\multirow{2}{*}{Verification} & \multirow{2}{*}{Prior} & \multicolumn{2}{c}{ThinkGeo} & EarthBench & TerraLogic \\
\cmidrule(lr){3-4} \cmidrule(lr){5-5} \cmidrule(lr){6-6}
& & Ans. & $\mathrm{Ans}_I$ & Acc. & AnsAcc \\
\midrule
\xmark & \xmark & 21.1 & 24.8 & 64.9 & 26.4 \\
\cmark & \xmark & 22.5 & 29.5 & 66.9 & 29.7 \\
\cmark & \cmark & \textbf{23.6} & \textbf{32.6} & \textbf{68.1} & \textbf{31.4} \\
\bottomrule
\end{tabular}
\end{table}

Table~\ref{tab:choice420} reports the remote sensing VQA results. All three metrics improve for all 11 backbone models, so all 33 comparisons show positive gains. Averaged over models, Overall, Perception, and Reasoning improve by 5.71, 5.84, and 5.45 percentage points, respectively.
Models with lower original accuracy usually have more room for improvement. Qwen3.5-122B has the lowest original Overall accuracy and obtains the largest gain of 13.57 percentage points. Its perception and reasoning accuracy improve by 14.29 and 12.14 percentage points, respectively. Qwen3.5-397B, Kimi K2.6, and Gemini-3.1-Pro start from high original accuracy, and their Overall gains are all no more than 2.14 percentage points.

Perception and reasoning gains differ across models. GPT-5 improves perception accuracy by 10.72 percentage points and reasoning accuracy by 1.42 percentage points. The corresponding gains of GPT-5.5 are 2.50 and 5.00 percentage points. Qwen3.5-122B gains substantially on both. These results show that the benefit of the framework varies with the backbone model and the task type.
With \ours{}, Gemini-3.1-Pro achieves the highest Overall accuracy of 89.05\%, GPT-5 achieves the highest Perception accuracy of 91.79\%, and GPT-5.5 achieves the highest Reasoning accuracy of 87.86\%. With the framework, some open-weight models also outperform closed-source models that answer directly. For example, Qwen3.5-397B reaches 85.95\% Overall, higher than the 80.48\% of GPT-5 with direct answering.

\begin{table}[t]
\centering
\caption{Performance and cost of tool verification strategies on EarthBench.}
\label{tab:verification_strategy}
\footnotesize
\setlength{\tabcolsep}{3pt}
\renewcommand{\arraystretch}{1.15}
\resizebox{\columnwidth}{!}{%
\begin{tabular}{@{}lccccc@{}}
\toprule
\multicolumn{1}{l}{\multirow{2}{*}{Method}}
& \multirow{2}{*}{\makecell{Acc.\\(\%)}}
& \multirow{2}{*}{\makecell{LLM\\calls}}
& \multirow{2}{*}{\makecell{Token\\usage}}
& \multicolumn{2}{c}{Time per question (s)} \\
\cmidrule(lr){5-6}
& & & & Mean & Median \\
\midrule
\base{}
& 64.9 & 7.117 & 90{,}454 & 460.37 & 321.99 \\
\base{} + CRITIC
& 66.9 & 8.871 & 148{,}562 & 610.45 & 505.76 \\
\base{} + Tools Fail
& 65.3 & 9.935 & 125{,}322 & 557.30 & 458.28 \\
\ours{}
& 68.1 & 8.367 & 107{,}860 & 477.08 & 334.46 \\
\bottomrule
\end{tabular}%
}
\end{table}

\subsection{Ablation Study}
\label{sec:ablation}
\textbf{Core component ablation.}
We use GPT-5 as the backbone and enable observation verification and the task--tool reliability prior in turn. When the prior is turned off, the verifier still checks the current image and process evidence but does not receive the historical reliability score. When both are turned off, the configuration becomes \base{} in Section~\ref{sec:setup}. All configurations use the same tool set and call budget.

\begin{table*}[t]
\centering
\caption{Statistics of observation verification and subsequent execution. The first four columns count tasks, and each ratio is conditioned on the previous column.}
\label{tab:detection}
\footnotesize
\setlength{\tabcolsep}{4pt}
\renewcommand{\arraystretch}{1.12}
\begin{tabular}{lccccc}
\toprule
Benchmark &
\shortstack{Tasks with a\\failed observation} &
\shortstack{Tasks with\\evidence requests} &
\shortstack{Tasks with an accepted\\observation after the request} &
\shortstack{Of which\\answered correctly} &
\shortstack{Rejected\\calls} \\
\midrule
EarthBench& 179/248 (72.2\%)& 177/179 (98.9\%)& 175/177 (98.9\%)& 104/175 (59.4\%)& 6 \\
ThinkGeo& 185/436 (42.4\%)& 168/185 (90.8\%)& 154/168 (91.7\%)& 77/154 (50.0\%)& 32 \\
TerraLogic& 205/545 (37.6\%)& 200/205 (97.6\%)& 191/200 (95.5\%)& 65/191 (34.0\%)& 14 \\
\midrule
Overall& 569/1,229 (46.3\%)& 545/569 (95.8\%)& 520/545 (95.4\%)& 246/520 (47.3\%)& 52 \\
\bottomrule
\end{tabular}
\end{table*}

The first row of Table~\ref{tab:ablation} is \base{}. Compared with \base{}, the full \ours{} improves the four metrics by 2.5, 7.8, 3.2, and 5.0 percentage points, and the gains cover all three benchmarks. Without the reliability prior, enabling observation verification alone also improves every metric. On ThinkGeo, Ans.\ and $\mathrm{Ans}_I$ improve by 1.4 and 4.7 percentage points. EarthBench and TerraLogic improve by 2.0 and 3.3 percentage points, respectively. The gain on $\mathrm{Ans}_I$ is clearly larger than that on Ans. Adding the reliability prior further improves the four metrics by 1.1, 3.1, 1.2, and 1.7 percentage points. This shows that a historical reference beyond the current process evidence is helpful. When a tool returns normally but its evidence is questionable, the prior shows how the configuration has performed on related tasks and helps the verifier decide whether more evidence is needed. The prior also improves $\mathrm{Ans}_I$ more than Ans. Both stages bring positive gains on all three benchmarks. Process evidence describes the current call, while the reliability prior provides a task-related historical reference. The two carry different information, and combining them outperforms verification without the prior.

\begin{table}[t]
\centering
\caption{Error propagation rate $\mathcal{R}_{\mathrm{prop}}$ of tool observations (\%). $\Delta$ is the change of \ours{} relative to \base{} in percentage points.}
\label{tab:error_propagation}
\footnotesize
\setlength{\tabcolsep}{5pt}
\renewcommand{\arraystretch}{1.08}
\begin{tabular}{lccc}
\toprule
Benchmark & \base{} & \ours{}& \shortstack{$\Delta$} \\
\midrule
EarthBench & 95.5 & 69.6 & $-25.9$ \\
ThinkGeo & 93.4 & 71.2 & $-22.2$ \\
TerraLogic & 96.8 & 75.5 & $-21.3$ \\
\bottomrule
\end{tabular}
\end{table}

\subsection{Analysis of Verification Strategies}
\label{sec:verification_strategy}

We compare task performance and inference cost across verification strategies.
CRITIC~\citep{gou2024critic} checks and revises an initial answer with tool feedback.
Tools Fail~\citep{sun2024toolsfail} judges whether a tool output is reliable based on the task context.
\ours{} verifies each tool observation before it enters the reasoning history.
The verification uses process evidence and the task--tool reliability prior. The system accepts the observation, requests additional evidence, or rejects it.

CRITIC and Tools Fail were not originally evaluated on remote sensing tasks.
We adapt both strategies to \base{}.
For CRITIC, the base agent first generates an initial answer. The model then calls remote sensing tools to check it and revises the answer based on the feedback. For Tools Fail, we use the checklist setting. The model judges tool outputs with the image, the task context, and the checklist. Accepted observations enter the reasoning history, and rejection feedback is returned to the planner. The planner then adjusts subsequent tool calls based on the feedback. Neither adaptation uses our reliability prior. All configurations use the same backbone, tool set, and test samples. The total tool-call budget is $H_{\max}=10$ for all configurations. Calls made for the initial solution, evidence requests, and corrections all count toward this budget.

Table~\ref{tab:verification_strategy} reports the performance and cost of each strategy. Compared with \base{}, \ours{} improves accuracy by 3.2 percentage points. The average number of LLM calls increases by 17.56\%. The average token usage increases by 19.24\%. The mean time per question increases by 3.63\%, and the median increases by 3.87\%. These times reflect the running cost under the preset time limit. Compared with the CRITIC adaptation, \ours{} is 1.2 percentage points more accurate. It also reduces the average token usage and time by 27.40\% and 21.85\%, respectively.
Compared with the Tools Fail adaptation, \ours{} is 2.8 percentage points more accurate. It also reduces the average token usage and time by 13.93\% and 14.39\%, respectively. Among the three verification configurations, \ours{} has the highest accuracy. It also has the fewest model calls, the lowest token usage, and the shortest mean time. Under the current evaluation setting, \ours{} achieves a more favorable trade-off between task performance and computational cost.

\subsection{Analysis of Observation Error Propagation}
\label{sec:error_analysis}

\textbf{Verification and evidence requests.} We track what happens after an observation fails verification. Table~\ref{tab:detection} reports the GPT-5 results. Given current evidence, the verifier requests more information or rejects the observation. The backbone model then uses feedback to adjust tool arguments or switch tools.

Of the 1,229 tasks, 569 (46.3\%) contain at least one observation that fails verification. Of these, 545 tasks request additional evidence, and 520 of them obtain at least one accepted observation after the request. On all three benchmarks, more than 90\% of tasks with failed observations request additional evidence. For an observation that fails verification, the system usually seeks more evidence rather than rejecting the observation directly. On EarthBench, 72.2\% of tasks contain at least one failed observation, higher than 42.4\% on ThinkGeo and 37.6\% on TerraLogic. This difference may come from how EarthBench tasks are organized. A task usually includes many files, and the agent needs to repeatedly inspect, select, and combine information across files. The execution produces more tool observations related to data selection and intermediate computation, and these observations may need further verification.

Of the 520 tasks with an accepted observation after the request, 246 (47.3\%) are answered correctly. This ratio varies across benchmarks: 59.4\% on EarthBench, 50.0\% on ThinkGeo, and 34.0\% on TerraLogic. The other 274 tasks still fail after obtaining accepted observations. Passing verification on a local observation does not mean that the whole task succeeds. After a verified observation is obtained, subsequent planning and multi-step reasoning still affect the final result. The system also rejects 52 observation calls. These rejected calls are not written to the verified history as accepted records. Sections~\ref{sec:case} and~\ref{sec:limitation} further analyze the related failure modes.

\textbf{Propagation of wrong tool observations.} Following Section~\ref{sec:empirical_analysis}, we compare the GPT-5 error propagation rates of \base{} and \ours{}.

\begin{figure*}[!t]
\centering
\includegraphics[width=\textwidth]{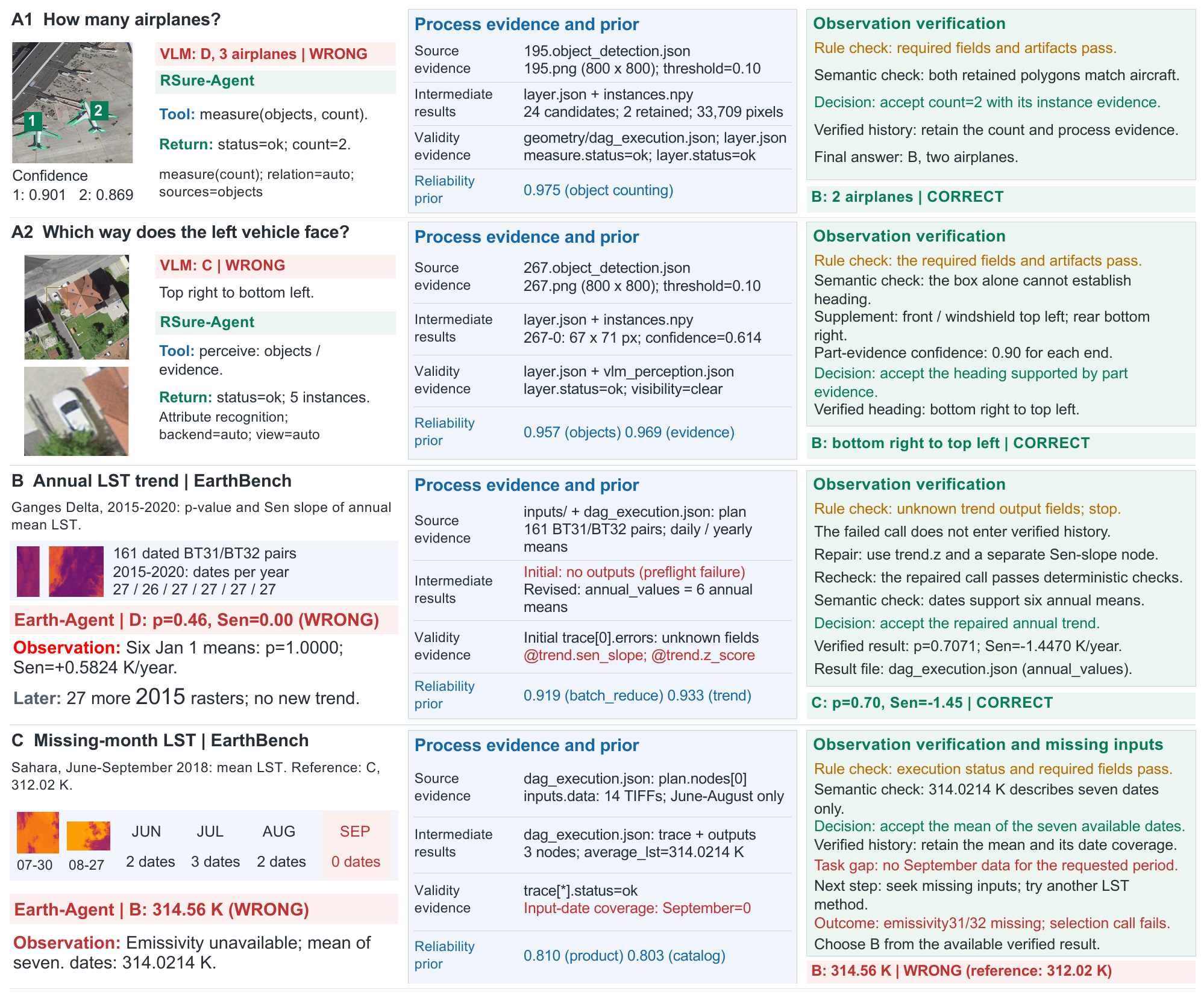}

\caption{Case comparison.
(A) The VLM miscounts aircraft (A1) and misjudges vehicle heading (A2).
\ours{} verifies observations with instance polygons and vehicle front and rear cues, obtaining correct answers.
(B) Earth-Agent uses single-day means for an annual trend test.
\ours{} aggregates data by year and corrects output fields using rule-check feedback.
The annual trend observation passes verification.
(C) \ours{} accepts the mean of seven June--August images, but September data are missing.
Missing inputs prevent an alternative computation.
Under the evaluation rule, it selects the incorrect option B.}
\label{fig:case_study}
\end{figure*}

Table~\ref{tab:error_propagation} reports the error propagation rates of \base{} and \ours{}.
In \base{}, this rate exceeds 93\% on all three benchmarks.
In other words, in almost all tasks with wrong tool observations, at least one wrong observation is adopted by subsequent reasoning, and the wrong final answer is consistent with it.
With the full \ours{} framework, the rate drops on all three benchmarks.
It falls from 95.5\% to 69.6\% on EarthBench, a drop of 25.9 percentage points; from 93.4\% to 71.2\% on ThinkGeo, a drop of 22.2 percentage points; and from 96.8\% to 75.5\% on TerraLogic, a drop of 21.3 percentage points.
Because verification may trigger follow-up calls and change the trajectory, the tasks with incorrect observations differ between the two methods.
Each rate is therefore computed over the tasks where that method produces incorrect observations.
These results show that the full framework substantially limits the effect of wrong tool observations on the final result.

\subsection{Case Study}
\label{sec:case}

Fig.~\ref{fig:case_study} compares \ours{} with direct VLM answering and Earth-Agent on benchmark tasks.

\textbf{Comparison with a VLM.}
In Fig.~\ref{fig:case_study}(A1), the VLM directly answers that there are three aircraft. \ours{} calls \texttt{measure} to count them, and the tool returns two aircraft with their instance polygons. The verifier matches the two polygons to the aircraft in the image one by one and confirms that both are supported by the image, so the count observation passes verification. The system writes the count and its instance evidence to the verified history, and the model gives the correct answer. In Fig.~\ref{fig:case_study}(A2), the VLM directly judges the heading of the left vehicle as top-right to bottom-left. \ours{} first locates the vehicle with \texttt{perceive(objects)}.
The detection box in the process evidence is $67\times71$ pixels, which is almost square, so the box alone cannot show where the vehicle front is. The backbone model then calls \texttt{perceive(evidence)} to obtain local features. The returned evidence shows that the front and windshield are at the upper left and the rear is at the lower right. The verifier checks the front and rear positions, confirms that the heading from bottom-right to top-left is supported by visual evidence, and accepts the observation. The model gives the correct answer. The two cases verify an instance count and a vehicle heading, respectively. Aircraft counting requires checking that the observed number matches the instances in the image. Vehicle heading requires checking that local features such as the front and rear support the direction. These checks link tool observations to specific instances and their visual features.

\textbf{Comparison with Earth-Agent.}
Fig.~\ref{fig:case_study}(B) asks for the trend of the annual mean land surface temperature (LST) over the Ganges Delta from 2015 to 2020. Earth-Agent uses only the image of January 1 of each year and inputs six single-day means to the trend test, which gives $p=1.0000$ and a Sen's slope of $+0.5824$~K/year. Its final answer D is also inconsistent with these tool results.
\ours{} uses all 161 images. It first computes the mean of each image and then aggregates the means by year,
so that each annual mean includes 26 to 27 images. The first trend call refers to the nonexistent fields \texttt{sen\_slope} and \texttt{z\_score},
and the deterministic rule check returns the specific error. The backbone model then corrects the output fields and computes the Sen's slope separately. It finally obtains $p=0.7071$ and $-1.4470$~K/year and selects the correct answer C. In this process, the source dates and image counts make the annual aggregation traceable, and the rule diagnosis helps the model locate and fix the call error.

Fig.~\ref{fig:case_study}(C) shows a case where a local observation is accepted but the evidence for the whole task is still incomplete. The question asks for the mean LST over the Sahara from June to September 2018, but the input contains only seven images from June to August. Earth-Agent computes 314.0214~K from the available data and directly selects option B (314.56~K). In contrast, \ours{} detects the missing September images before the first tool call. It computes the same 314.0214~K from the available data. The observation passes the rule check, and the verifier confirms that the value matches the mean of the seven input images. The system accepts the observation and records its data source and temporal coverage in the verified history. The observation can be used in subsequent reasoning, but it is not enough to determine the full June--September mean, and the coverage gap remains. The computed value differs from the closest option by about 0.54~K, so the backbone model tries another LST algorithm to obtain more evidence. However, this algorithm needs \texttt{emissivity31/32}, which is also missing. The computation cannot run, and the later option-selection call also fails. Following the evaluation rule, the model selects B, the option closest to the computed value, which differs from the reference answer C (312.02~K). This case shows that observation verification is limited by tool capabilities. Verification can identify evidence gaps and guide the backbone model to adjust its calls. Whether the gaps can be filled, however, depends on tool data access and computing capabilities. Verification alone cannot compensate for insufficient tool capabilities.

\begin{table}[t]
\centering
\caption{Tool-call efficiency on EarthBench. -- means not reported.}
\label{tab:efficiency}
\footnotesize
\setlength{\tabcolsep}{4pt}
\renewcommand{\arraystretch}{1.08}
\resizebox{\linewidth}{!}{
\begin{tabular}{llcccc}
\toprule
Method & Backbone & Spectrum & Products & RGB & Overall \\
\midrule
Earth-Agent & GPT-4o& 3.7736 & 1.1800 & 0.8779 & 2.1579 \\
Earth-Agent & GPT-5& 3.5510 & 1.7154 & 1.5312 & 2.4146 \\
HiRS-Agent & Qwen3-4B& -- & -- & -- & 2.5059 \\
HiRS-Agent & Qwen3-8B& -- & -- & -- & 2.5248 \\
\midrule
\ours{} & GPT-5& 2.2697 & 1.6103 & 1.2502 & 1.7891 \\
\bottomrule
\end{tabular}}
\end{table}

\subsection{Efficiency Analysis}
\label{sec:efficiency}

We use the Efficiency metric of EarthBench to measure tool-call cost. For each task, the metric divides the actual call count by the expert reference call count, then averages these ratios.

Table~\ref{tab:efficiency} reports the tool-call efficiency on EarthBench. The overall tool-call ratio of \ours{} is 1.7891, lower than the 2.4146 of Earth-Agent with the same GPT-5 backbone, a relative reduction of 25.9\%. All follow-up calls triggered by verification are included in the total count. Even with these calls, the overall ratio of \ours{} is still lower than that of Earth-Agent with GPT-5.
The metric counts tool calls only and excludes the extra inference cost of the verifier. On every task subset, namely Spectrum, Products, and RGB, \ours{} has a lower tool-call ratio than Earth-Agent with GPT-5. The ratio drops from 3.5510 to 2.2697 on Spectrum, a relative reduction of 36.1\%. It drops from 1.7154 to 1.6103 on Products, a relative reduction of 6.1\%, and from 1.5312 to 1.2502 on RGB, a relative reduction of 18.4\%. Together with Table~\ref{tab:main_results}, these results show that \ours{} improves task accuracy and reduces the tool-call ratio relative to Earth-Agent with GPT-5. The Overall ratios of all configurations are still above 1, so there is room to further reduce tool calls relative to the expert reference.

\subsection{Limitations}
\label{sec:limitation}

The effect of \ours{} depends on the available data and the capabilities of existing tools. When the observations needed for a task are missing from the input data and no existing tool can obtain them, further verification cannot produce new valid observations. For example, the period-mean case in Section~\ref{sec:case} lacks September observations, which no tool can fill. Additionally, the verifiable observation protocol depends on tools that provide process evidence for their observations. The perception, measurement, and execution tools in this work return source evidence, intermediate results, and validity evidence, allowing the verifier to check whether observations are supported. For tools that return only final results without full process evidence, less information is available for verification. Finally, the reliability of tool observations is not the only factor that determines the task outcome. In Table~\ref{tab:detection}, some tasks still fail after obtaining accepted observations, and in Table~\ref{tab:error_propagation}, the error propagation rate of \ours{} is still 69.6\% to 75.5\%. These results show that more reliable tool observations reduce the effect of incorrect observations on subsequent reasoning, but the final task performance depends on both the quality of tool observations and the subsequent reasoning process.

\section{Conclusion}
\label{sec:conclusion}

In this paper, we present \ours{}, a framework for limiting the effect of incorrect tool observations on subsequent reasoning in remote sensing agents. Motivated by our analysis of 1,229 execution trajectories, the framework combines process evidence from a verifiable observation protocol with a task--tool reliability prior. The prior provides historical information about tool reliability across task types to support verification. Together, these sources guide decisions to accept observations, request additional evidence, or reject observations. Experiments on EarthBench, ThinkGeo, and TerraLogic show competitive task performance. Across these benchmarks, \ours{} reduces the error propagation rate by at least 21 percentage points. On EarthBench, it improves task accuracy and reduces the tool-call ratio by 25.9\% relative to Earth-Agent. On CHOICE-420, it improves overall accuracy on average across 11 backbone models. These results support verifying tool observations before accepting them as evidence for subsequent reasoning. Future work focuses on using verification feedback to correct subsequent reasoning and extending observation verification to a broader, more diverse range of Earth observation tasks.

%\section*{Declaration of generative AI and AI-assisted technologies in the manuscript preparation process}

%During the preparation of this work, the authors used ChatGPT (OpenAI) to improve the language, clarity, and organization of the manuscript. The authors reviewed and edited the AI-assisted content and take full responsibility for the content of the published article.

\appendix
\counterwithin*{table}{section}
\counterwithin*{figure}{section}
\renewcommand{\thetable}{\Alph{section}.\arabic{table}}
\renewcommand{\thefigure}{\Alph{section}.\arabic{figure}}
\renewcommand{\theHtable}{\Alph{section}.\arabic{table}}
\renewcommand{\theHfigure}{\Alph{section}.\arabic{figure}}

\section{Tool Descriptions and Protocol Parameters}
\label{app:tools}

This appendix lists the tool descriptions and key argument fields visible to the model. In the code, the tools are registered as \texttt{rs\_perceive}, \texttt{rs\_measure}, \texttt{rs\_execute}, and \texttt{rs\_retrieve}. Braced variables in the prompts, such as \texttt{\{objective\}}, are filled in at run time.

\subsection{Top-Level Tool Descriptions}

\begin{promptbox}{\texttt{perceive} tool description}
Unified visual evidence for remote sensing. Use mode=evidence for image/figure VQA facts, document only for literal transcription of text/axes/legends, and scene for taxonomy-constrained classification. Use objects for instance boxes, regions for semantic masks, change for ordered before/after images, or measurement for two landmarks and a visible scale bar. For scene, put semantic land-use labels in label\_space and expect one exact label per source image. Never use the final answer letters as scene classes. Use objects or regions when later geometry must be computed; evidence does not choose the final answer. One scene backend is selected per image set and reused. Numerical measurements and coordinate-based spatial relations must be computed deterministically. Qualitative heading judgments may use visible front and rear features.
\end{promptbox}

\begin{promptbox}{\texttt{measure} tool description}
Produce one numerical spatial observation from explicitly declared image sources and a typed geometry specification. Each source states whether it needs instances, semantic regions, an ordered T1/T2 transition, or a caller-declared rectangular coordinate extent. The spec states the metric, labels, relation, distance, denominator, unit, and GSD. Sources must name raw visual evidence only. Use objects for discrete instances, regions for continuous or linear surfaces, change for a T1/T2 transition, and extent only for explicit frame coordinates. Put buffers, overlap, ratios, and zones in spec. The tool does not infer these from wording, read candidate answers, or use benchmark metadata. Prefer it to a hand-written spatial DAG when the required result is a count, area, fraction, percentage, distance, overlap, or boolean relation over visible imagery.
\end{promptbox}

\begin{promptbox}{\texttt{execute} tool description}
Compile and execute a deterministic DAG for raster formulas, statistics, time series, arithmetic, and spatial relations over SpatialLayer artifacts. Connect ordinary stages with inputs.data=@node. Prefer this tool whenever the requested answer is a number, area, distance, percentage, trend, threshold result, or spatial count. Return the measured value and compare it directly with answer choices when provided. Unknown operators and invalid units fail closed.
\end{promptbox}

\begin{promptbox}{\texttt{retrieve} tool description}
Retrieve a small set of trusted remote-sensing capability, algorithm-recipe, or domain-knowledge cards. Use it for formulas, sensor/physics concepts, or to identify the abstract workflow. Retrieval provides context, not a final numeric answer; execute calculations with rs\_execute.
\end{promptbox}

\begin{table}[t]
\centering
\footnotesize
\caption{Key arguments of \texttt{perceive}.}
\label{tab:perceive_params}
\setlength{\tabcolsep}{3pt}
\begin{tabularx}{\columnwidth}{@{}>{\raggedright\arraybackslash}p{2.6cm}>{\raggedright\arraybackslash}X@{}}
\toprule
Argument & Description visible to the model (summary) \\
\midrule
\texttt{image\_paths} &
Input image paths in their semantic order. \\
\texttt{objective} &
Exact visual fact or layer required by the task. \\
\texttt{mode} &
One of seven perception modes: evidence / document / scene / objects / regions / change / measurement. \\
\texttt{label\_space} &
Exact semantic classes for mode=scene, separate from the outer multiple-choice response labels. \\
\texttt{target\_groups} &
Semantic target id to one or more open-vocabulary prompts; prompts in one group are fused as one class. \\
\texttt{landmarks} &
Exactly two visible landmark descriptions, required only for measurement mode. \\
\texttt{view} &
tiles adds non-overlapping detail views for large images. \\
\texttt{gsd} &
One ground sample distance for all images, or one meters-per-pixel value per image. \\
\texttt{quality\_policy} &
strict enables an opt-in confidence/geometry gate and a retry when a required target group is missing. \\
\bottomrule
\end{tabularx}
\end{table}

\begin{table}[t]
\centering
\footnotesize
\caption{Key arguments of \texttt{measure}.}
\label{tab:measure_params}
\setlength{\tabcolsep}{3pt}
\begin{tabularx}{\columnwidth}{@{}>{\raggedright\arraybackslash}p{3.5cm}>{\raggedright\arraybackslash}X@{}}
\toprule
Argument & Description visible to the model (summary) \\
\midrule
\texttt{sources} &
Explicit perception sources. Declare the needed geometry instead of asking the tool to infer it from question wording. \\
\texttt{spec} &
Typed metric, subject/reference labels, spatial relation, optional distance, denominator, and output unit. \\
\texttt{source.mode} &
objects / regions / change / extent; derived spatial relations go only in spec. \\
\texttt{source.gsd} &
Meters per pixel for each source image. Required when this layer is used with meter or square-meter geometry. \\
\texttt{source.backend} &
Use local for measurable geometry. VLM geometry is approximate and may be rejected by the measurement contract. \\
\texttt{spec.distance\_statistic} &
minimum is the closest pair; mean and maximum aggregate one edge distance per subject instance or component. \\
\bottomrule
\end{tabularx}
\end{table}
\subsection{Key Argument Fields}

Tables~\ref{tab:perceive_params} and~\ref{tab:measure_params} summarize the key arguments.

Perception tools that produce geometry layers also return the following contract.

\begin{promptbox}{Geometry contract (geometry\_contract)}
Exact geometry is stored in layer\_paths. Use rs\_measure or rs\_execute with spatial.query to compute distances, areas, and coordinate-based spatial relations. Do not estimate these quantities or relations from the summary.
\end{promptbox}

\subsection{Standard Products of \texttt{execute}}
\label{app:products}

\texttt{execute} includes five LST retrieval products: \texttt{single\_\allowbreak channel\_\allowbreak lst}, \texttt{multi\_\allowbreak channel\_\allowbreak lst}, \texttt{split\_\allowbreak window\_\allowbreak lst}, \texttt{ttm\_\allowbreak lst}, and \texttt{tes}. It also includes five other products: \texttt{normalized\_\allowbreak difference}, \texttt{pwv\_\allowbreak band\_\allowbreak ratio}, \texttt{tvdi}, \texttt{day\_\allowbreak night\_\allowbreak temperature}, and \texttt{apparent\_\allowbreak thermal\_\allowbreak inertia}. A product is declared through \texttt{asset.\allowbreak catalog}, mapped to each observation by \texttt{raster.\allowbreak product}, and aggregated by \texttt{raster.\allowbreak batch\_\allowbreak reduce} into a timestamped scalar series.

\section{Agent and Verification Prompts}
\label{app:prompts}

\subsection{Main System Prompt}
\label{app:system_prompt}

\begin{promptbox}{Main system prompt (SYSTEM\_PROMPT)}
You are RSure, a remote-sensing analysis agent with four tools.

- rs\_retrieve supplies trusted concepts and standard workflows.
- rs\_perceive extracts option-neutral visual evidence, documents, object layers, region layers, change layers, or a scale-bar measurement.
- rs\_measure turns explicitly declared visual sources, spatial relations, units, and GSD into one deterministic numeric observation.
- rs\_execute runs typed DAGs for products, raster math, statistics, time series, and spatial relations.

Plan from evidence, never benchmark names or shortcuts. Retrieve only missing knowledge. Ground pixel claims in perception. For spatial measurements use rs\_measure with declared sources, relation, units, and GSD. Use objects for instances, regions for surfaces, change for ordered T1/T2, and extent for stated rectangles. Put derived relations in spec. Execute numeric, temporal, and coordinate-based spatial computations. Ground qualitative heading judgments in visible front and rear features.

EO algorithms require physical product inputs. Prefer one complete DAG. Use inputs.data=@node; fields only for non-primary results. raster.summarize takes one raster; product collections use raster.batch\_reduce before temporal stages. Filter phases in asset.catalog params.phases; bind ambiguous bXX roles with role\_bands; pass date bounds in params.start/end. Use masks only for spatial validity. threshold\_fraction uses params.threshold/comparison.

For scene classification, declare taxonomy in label\_space, never final A/B/C/D labels. Use one exact class per source image and repair insufficient output. Compare successful measurements with choices when provided; use decision.select for numeric candidates. Failed or low-quality evidence never means absence. Do not repeat successful calls, invent constants, guess measurable values, or artifact paths. Return only requested concise format.
\end{promptbox}

\subsection{Online Verification Prompt}
\label{app:verify_prompt}

The verifier $M_{\mathrm{eval}}$ receives the inputs in Eq.~\eqref{eq:sem_check} with the prompt below.

\begin{promptbox}{Online verification prompt}
You are an independent observation verifier for a remote-sensing agent. Decide whether the current tool observation is sufficiently supported.

TASK TYPE: \{task\_type\}
QUESTION: \{question\}
TOOL ACTION: \{tool\_action\}
TOOL OBSERVATION: \{observation\}
PROCESS EVIDENCE: \{evidence\}
RELIABILITY PRIOR: \{prior\}

Check:
1. What should be checked for the current task?
2. Does the image support the observation?
3. Does the process evidence support the observation?
4. Is additional evidence needed?
5. Use the reliability prior only as historical reference. A low prior requires stricter evidence; a high prior never waives evidence. If the prior is None, decide from the remaining inputs only.

Decide exactly one:
- accept: the evidence sufficiently supports the observation.
- request\_evidence: additional evidence is needed to determine whether the current observation is correct or applicable.
- reject: the evidence shows the observation is wrong or inapplicable; state the reason.
\end{promptbox}

\subsection{Final Answer Synthesis Prompt}
\label{app:final_prompt}

The final answer is generated in a new context with the following prompt.

\begin{promptbox}{Final answer synthesis prompt}
Answer the original task using only the verified observations below. Ignore rejected or unverified results. If the verified observations are insufficient for the requested measurement, follow the task-specific answer requirement. When an explanatory response is allowed, state what evidence is missing. Do not invent missing measurements. Return only the required final answer.

TASK: \{task\}
ANSWER REQUIREMENT: \{response\_contract\}
VERIFIED OBSERVATIONS: \{verified\_history\}
\end{promptbox}

The answer requirement $c$ comes from the benchmark adapter (\ref{app:task_templates}), and numerical answers carry the units required by the task. 

\subsection{VLM Perception Prompts}
\label{app:vlm_prompts}

Each perception mode sends a common prefix and mode-specific instructions to the VLM.

\begin{promptbox}{Common prefix}
You are a remote-sensing perception component. Use only visible evidence. Return one strict JSON object, without markdown. Modality hint: \{modality\}. GSD: \{gsd\} m/px. Objective: \{objective\}
\end{promptbox}

The mode-specific instructions follow. The \texttt{evidence} mode stays option-neutral.

\begin{promptbox}{evidence mode}
Provided views in order: \{view\_descriptions\}. Detail tiles from one source do not overlap; use the overview for context and count each tile exactly once.
Candidate options: \{options\}

Extract visible evidence without choosing a final answer. Candidate concepts only define which visible distinctions matter; do not compare, rank, support, contradict, or recommend candidates. Do not return answer, answer\_label, or a preferred option.

For workflow diagrams, separate the purpose of the full input-output graph from the names of internal steps. For plots, identify the axis scale first. Anchor at least two visible tick values to their positions, and interpolate peaks or crossings in the transformed axis coordinates. Never estimate values from apparent spacing alone. Report plot calibration in measurements with the scale and at least two \{tick\_value, normalized\_position\} anchors. Include each important feature's normalized\_position and interpolated\_value.

Return keys evidence\_kind (scene/diagram/plot/map/document/geometry/change), observations, visible\_text, measurements, spatial\_relations, quality, missing\_evidence, uncertainty. Each observation has claim, view\_index, optional region, and confidence. quality has visibility, readability, and issues. quality.visibility must be clear, limited, or unusable; quality.readability must be clear, limited, or unreadable.
\end{promptbox}

\begin{promptbox}{document mode}
Transcribe only visible text, axes, legends, labels, panels, and their spatial relations. Do not identify an unlabeled subject, interpret its meaning, compare candidates, or answer the objective. Return keys text, structure, confidence.
\end{promptbox}

\begin{promptbox}{measurement mode}
Displayed views in order: \{view\_descriptions\}.
Landmark A: \{landmark\_A\}
Landmark B: \{landmark\_B\}

Locate only those two visible landmarks and one visible scale bar from the same source image. Do not infer an answer, choose a candidate, estimate a final ground distance, or use information not visible in the image. Coordinates must be in pixels of the corresponding displayed view, with origin at upper-left. For a labeled or arrowed feature, locate the indicated physical marker or arrow tip, never the label glyph itself. For a scale bar, locate the endpoints of the physical bar line or tick marks, never its text label or surrounding box.

Return exactly \{\allowbreak measurement:\allowbreak \{\allowbreak view\_index, \allowbreak point\_a: \{label, xy, view\_index, uncertainty\_px\}, \allowbreak point\_b: \{label, xy, view\_index, uncertainty\_px\}, \allowbreak scale\_bar: \{start\_xy, end\_xy, view\_index, ground\_length, unit, uncertainty\_px, ground\_length\_uncertainty\}, \allowbreak confidence\}\}.

A point and the scale bar may use different displayed detail views of the same source image. ground\_length is the numeric value printed by the scale bar, and unit is its separate visible unit text.
\end{promptbox}

\begin{promptbox}{scene mode}
Candidate scene classes: \{options\}
Displayed source images in order: \{view\_descriptions\}.

Classify the overall scene of every source image using visible land use, geometry, and context. Return exactly one strict JSON object with keys predictions and evidence. predictions must contain exactly one object for each source image, with no ranked alternatives. Each object has the following fields: \{image\_index: 1-based source image index, class\_name: one exact candidate scene class, confidence: number from 0 to 1, evidence: short visible justification\}. Every image\_index from 1 through the number of source images must appear exactly once. Do not return final answer letters or a label outside Candidate scene classes.
\end{promptbox}

\begin{promptbox}{objects / regions mode}
Target groups: \{target\_groups\}
Displayed image views in order: \{view\_descriptions\}.

When detail tiles are present, use overviews only for context and report geometry from the non-overlapping detail tiles, so each target is returned once. Return key \{mode\}, a list. Each item must contain view\_index (1-based), target\_group\_id, label, confidence, and either bbox\_xyxy=[x1,y1,x2,y2] or polygon\_xy=[[x,y],...] in displayed-image pixels.
\end{promptbox}

\begin{promptbox}{change mode}
Compare the images in order. Report changed\_regions on the corresponding displayed view. Each changed region must contain view\_index, target\_group\_id, label, confidence, and either bbox\_xyxy=[x1,y1,x2,y2] or polygon\_xy=[[x,y],...]. Return keys changed\_regions, evidence, confidence, uncertainty.
\end{promptbox}

The local zero-shot scene backend ensembles four templates, \texttt{a satellite image of \{label\}}, \texttt{an aerial photograph of \{label\}}, \texttt{a remote sensing scene of \{label\}}, and \texttt{a top-down remote sensing view of \{label\}}, for \texttt{\{label\}} in the declared \texttt{label\_space}.

\subsection{Task Templates}
\label{app:task_templates}

Each benchmark adapter turns an official sample into the task text below. For benchmarks with candidate answers, the model must choose one candidate from the available evidence even if an information gap remains. This rule follows the benchmark evaluation setting and is also part of the answer requirement $c$, as shown in the EarthBench template.

\begin{promptbox}{EarthBench task template}
Solve this remote-sensing data-analysis task.
Clean data directory: \{clean\_dir\}
Official input files (\{n\}):
\{inventory\}

Question: \{question\}
\{options\}
Return the final result in the format \{answer\_format\}.

When candidate answers are provided for this benchmark, return exactly one candidate answer in the required format.
\end{promptbox}

\begin{promptbox}{ThinkGeo task template}
Solve this remote-sensing task from the supplied images.
Images:
\{images\}

Task: \{question\}
Return a concise concrete answer containing every requested value, direction, count, and unit.
\end{promptbox}

\begin{promptbox}{TerraLogic task template}
Solve this remote-sensing spatial-analysis task without candidate options.
Images:
\{image\_text\}
Question: \{question\}
Return only the requested numeric result and unit as \texttt{<Answer>value unit</Answer>}.
\end{promptbox}

\texttt{retrieve} performs semantic retrieval over 29 local cards: 3 capability-routing cards, 13 algorithm recipes, and 13 domain-knowledge cards (Table~\ref{tab:cards}). For example, \texttt{recipe.temporal\_trend} defines the per-image aggregation and Theil--Sen trend computation used in Fig.~\ref{fig:case_study}(B), and \texttt{recipe.thermal\_raster\_products} lists the physical inputs of each LST product, including the \texttt{emissivity31/32} missing in Fig.~\ref{fig:case_study}(C). \texttt{knowledge.landsat\_qa\_pixel} defines the QA\_PIXEL validity bit masks used as validity evidence. All cards will be released with the code.

\section{Task Types of CHOICE-420}
\label{app:choice_tasks}

Table~\ref{tab:choice_task_types} lists the 21 task types of CHOICE-420. Following CHOICE~\citep{li2025choice}, perception covers image-level comprehension, single-instance identification, and cross-instance discernment, and reasoning covers attribute, assessment, and common-sense reasoning. All tasks are multiple-choice. For example, Image Caption selects a candidate caption instead of generating one, and Object Localization selects a position description instead of outputting box coordinates.

\begin{table*}[!t]
\centering
\footnotesize
\caption{Task types of CHOICE-420 (20 samples per type).}
\label{tab:choice_task_types}
\setlength{\tabcolsep}{5pt}
\renewcommand{\arraystretch}{1.08}
\begin{tabularx}{\textwidth}{@{}
  >{\raggedright\arraybackslash}p{1.6cm}
  >{\raggedright\arraybackslash}p{3.1cm}
  >{\raggedright\arraybackslash}X
  >{\raggedright\arraybackslash}p{2.7cm}@{}}
\toprule
Category & Task Type & Definition & Answer Form \\
\midrule
Perception & Image Modality & Identify the sensor type or imaging mode of the image & Image modality \\
& Image Quality & Judge or compare image quality & Quality judgment \\
& Map Recognition & Identify the region from the outline or content of a map & Region \\
& Scene Classification & Identify the overall scene category of the image & Scene category \\
& Image Caption & Judge whether a candidate caption matches the overall image content & Matching caption \\
& Landmark Recognition & Identify a specific landmark in the image & Landmark name \\
& Object Counting & Count the objects of a given category & Object count \\
& Object Localization & Determine the position of a target in the image & Position description \\
& Object Presence & Determine whether a given target exists & Presence judgment \\
& Attribute Recognition & Identify the category or visible attributes of a given instance & Attribute \\
& Hallucination Detection & Judge whether a question or description contains content not supported by the image & Consistency judgment \\
& Attribute Comparison & Compare the visible attributes of multiple instances & Comparison result \\
& Spatial Relationship & Determine the relative positions of targets & Spatial relationship \\
& Change Detection & Identify changes in the same area at different times & Change description \\
\midrule
Reasoning & Time Property & Infer temporal attributes such as the season from the image & Time property \\
& Physical Property & Infer physical attributes such as the height of a target & Value \\
& Environmental Assessment & Estimate environmental indicators or conditions from the image & Indicator level \\
& Resource Assessment & Estimate resource indicators such as population from the image & Estimate \\
& Disaster Discrimination & Determine the disaster type from the image & Disaster type \\
& Geospatial Determination & Determine the city or country from cues such as landmarks & Location \\
& Situation Inference & Infer related activities or objects from the scene and common sense & Situation judgment \\
\bottomrule
\end{tabularx}
\end{table*}

\section{Retrieval Card Library}
\label{app:cards}

Table~\ref{tab:cards} summarizes the retrieval card library used by the \texttt{retrieve} tool. The tool performs semantic retrieval over 29 local cards, comprising 3 capability-routing cards, 13 algorithm recipes, and 13 domain-knowledge cards. Specifically, capability-routing cards guide tool selection, algorithm recipes define standard computation workflows and input requirements, and domain-knowledge cards provide sensor parameters and physical conventions. For example, \texttt{recipe.temporal\_trend} defines the per-image aggregation and Theil--Sen trend computation used in Fig.~\ref{fig:case_study}(B), and \texttt{recipe.thermal\_raster\_products} lists the physical inputs of each LST product, including the \texttt{emissivity31/32} bands missing in Fig.~\ref{fig:case_study}(C). In addition, \texttt{knowledge.landsat\_qa\_pixel} defines the QA\_PIXEL validity bit masks used as validity evidence. All retrieval cards will be released with the source code.

\begin{table}[!t]
\centering
\scriptsize 
\caption{Retrieval card index.}
\label{tab:cards}
\setlength{\tabcolsep}{3pt}
\renewcommand{\arraystretch}{0.95}
\begin{tabularx}{\columnwidth}{@{}lX@{}}
\toprule
id & title \\
\midrule
\multicolumn{2}{@{}l}{capability (3 cards)} \\
\hspace{4pt}cap.visual & Visual perception \\
\hspace{4pt}cap.compute & Deterministic computation \\
\hspace{4pt}cap.dag\_contract & Deterministic DAG contract \\
\multicolumn{2}{@{}l}{recipe (13 cards)} \\
\hspace{4pt}recipe.normalized\_difference & Normalized difference \\
\hspace{4pt}recipe.threshold\_ratio & Threshold fraction \\
\hspace{4pt}recipe.conditional\_raster\_statistics & Conditional raster statistics \\
\hspace{4pt}recipe.thermal\_raster\_products & Thermal raster product retrieval \\
\hspace{4pt}recipe.pwv\_band\_ratio & Near-infrared PWV band ratio \\
\hspace{4pt}recipe.tvdi & Temperature Vegetation Dryness Index \\
\hspace{4pt}recipe.temporal\_trend & Temporal aggregation and trend \\
\hspace{4pt}recipe.spatial\_relation & Spatial relation over layers \\
\hspace{4pt}recipe.change & Multi-temporal change \\
\hspace{4pt}recipe.raster\_collection\_aggregation & Aligned raster collection aggregation \\
\hspace{4pt}recipe.series\_distribution\_and\_sen & Distribution statistics and robust trend \\
\hspace{4pt}recipe.day\_night\_temperature & Paired day and night temperature \\
\hspace{4pt}recipe.apparent\_thermal\_inertia & Apparent thermal inertia \\
\multicolumn{2}{@{}l}{knowledge (13 cards)} \\
\hspace{4pt}knowledge.raster\_scaling & Raster storage scale and physical values \\
\hspace{4pt}knowledge.sar & SAR interpretation \\
\hspace{4pt}knowledge.hyperspectral & Spectral and hyperspectral reasoning \\
\hspace{4pt}knowledge.photogrammetry & Photogrammetry \\
\hspace{4pt}knowledge.gnss & GNSS \\
\hspace{4pt}knowledge.lidar\_dem & LiDAR and elevation \\
\hspace{4pt}knowledge.thermal & Thermal remote sensing \\
\hspace{4pt}knowledge.atmospheric\_spectroscopy & Atmospheric transmission and spectroscopy \\
\hspace{4pt}knowledge.cartography & Cartography and map projections \\
\hspace{4pt}knowledge.dem\_derivatives & DEM slope and aspect \\
\hspace{4pt}knowledge.image\_quality & Remote-sensing image quality \\
\hspace{4pt}knowledge.remote\_sensing\_resolution & Remote-sensing resolution and sampling \\
\hspace{4pt}knowledge.landsat\_qa\_pixel & Landsat QA\_PIXEL validity bits \\
\bottomrule
\end{tabularx}
\end{table}

\section{Task--Tool Reliability Prior}
\label{app:reliability_prior}

The prior matrices are built from the 1,200 calibration tasks in Section~\ref{sec:prior} and use the 21 task types in Appendix~\ref{app:choice_tasks}. The coarse and fine matrices have 37 and 88 configuration columns, of which 27 and 18 have non-empty scores. Each level is filtered by its own call counts and keeps only entries with $n(z,v)\ge n_{\min}=8$. A dash (--) indicates an unavailable prior, not a zero score. Since this filtering is done offline, the exported matrices contain only scores, and online queries apply Eq.~\eqref{eq:hierarchical_prior} to kept entries.

\subsection{Coarse Reliability Matrix}

Coarse configurations are defined by \texttt{mode} for perception, \texttt{metric} for measurement, and \texttt{op} for execution. Table~\ref{tab:reliability_matrix_coarse_part1} lists the perception modes and selected measurement metrics and execution operators, grouped by tool. Each score belongs to one task and configuration and should be queried with the full configuration identifier.

\subsection{Examples of the Fine Reliability Matrix}

Fine perception configurations are defined by mode, backend, and view, where \texttt{auto} denotes automatic selection. Fine measurement configurations are defined by metric, spatial relation, and input layer type, with the original export labels. Table~\ref{tab:reliability_matrix_fine_perceive} shows perception and measurement examples in panels (a) and (b), respectively. The fine export has no execution columns. Each level computes its scores from its own call counts, so coarse scores are not averages of fine scores.

\begin{table*}[!t]
\centering
\footnotesize
\setlength{\tabcolsep}{2.5pt}
\renewcommand{\arraystretch}{1.08}
\caption{Coarse reliability prior grouped by tool.}
\label{tab:reliability_matrix_coarse_part1}

\begin{tabular*}{\textwidth}{@{\extracolsep{\fill}}l*{14}{c}@{}}
\toprule
& \multicolumn{7}{c}{Perception}
& \multicolumn{4}{c}{Measurement}
& \multicolumn{3}{c}{Execution} \\
\cmidrule(lr){2-8}
\cmidrule(lr){9-12}
\cmidrule(lr){13-15}
Task type &
change & doc. & evid. & meas. &
objects & regions & scene &
area & boolean & count & distance &
\makecell{asset.\\catalog} &
\makecell{raster.\\product} &
\makecell{raster.\\summarize} \\
\midrule
Attribute Comparison
& -- & -- & 0.970 & -- & 0.909 & -- & --
& -- & -- & -- & -- & -- & -- & -- \\
Change Detection
& 0.321 & -- & 0.947 & -- & 0.830 & -- & --
& 0.833 & -- & 0.964 & -- & -- & -- & -- \\
Spatial Relationship
& -- & -- & 0.988 & -- & 0.880 & 0.818 & --
& 0.857 & -- & 0.951 & 0.860 & -- & -- & -- \\
Image Caption
& -- & -- & 0.923 & -- & -- & -- & 0.909
& -- & -- & -- & -- & -- & -- & -- \\
Image Modality
& -- & -- & 0.966 & -- & -- & -- & --
& -- & -- & -- & -- & -- & -- & -- \\
Image Quality
& -- & -- & 0.964 & -- & -- & -- & --
& -- & -- & -- & -- & -- & -- & -- \\
Map Recognition
& -- & -- & 0.893 & -- & -- & -- & --
& -- & -- & -- & -- & -- & -- & -- \\
Scene Classification
& -- & -- & -- & -- & -- & -- & 0.955
& -- & -- & -- & -- & -- & -- & -- \\
Attribute Recognition
& -- & -- & 0.974 & -- & 0.958 & -- & --
& -- & -- & -- & -- & -- & -- & -- \\
Hallucination Detection
& -- & -- & 0.909 & -- & 0.455 & -- & --
& -- & -- & -- & -- & -- & -- & -- \\
Landmark Recognition
& -- & -- & 0.963 & -- & -- & -- & --
& -- & -- & -- & -- & -- & -- & -- \\
Object Counting
& -- & -- & 0.971 & -- & 0.926 & 0.750 & --
& 0.714 & -- & 0.968 & -- & -- & -- & -- \\
Object Localization
& -- & -- & 0.971 & -- & 0.931 & 0.600 & --
& -- & -- & 0.933 & 0.818 & -- & -- & 0.971 \\
Object Presence
& -- & -- & -- & -- & 0.800 & -- & --
& -- & -- & 0.867 & -- & -- & -- & -- \\
Environmental Assessment
& -- & 0.875 & 0.885 & -- & 0.852 & -- & 0.913
& -- & -- & -- & -- & 0.803 & 0.810 & 0.973 \\
Resource Assessment
& -- & -- & 0.964 & -- & 0.571 & -- & --
& -- & -- & 0.727 & -- & -- & -- & 0.900 \\
Physical Property
& -- & -- & 0.956 & 0.200 & 0.872 & 0.750 & --
& 0.686 & -- & -- & 0.833 & -- & -- & -- \\
Time Property
& -- & -- & 0.963 & -- & -- & -- & --
& -- & -- & -- & -- & -- & -- & -- \\
Disaster Discrimination
& 0.346 & -- & 0.936 & -- & 0.865 & 0.938 & --
& 0.800 & 0.900 & 0.897 & 0.944 & -- & -- & -- \\
Geospatial Determination
& -- & -- & 0.955 & -- & -- & -- & --
& -- & -- & -- & -- & -- & -- & -- \\
Situation Inference
& -- & -- & 0.973 & -- & 0.733 & -- & --
& -- & -- & -- & -- & -- & -- & -- \\
\bottomrule
\end{tabular*}

\par\smallskip
{\scriptsize\raggedright
doc. = \texttt{document};
evid. = \texttt{evidence};
meas. = \texttt{measurement}.
\par}

\vspace{10pt}
\caption{Fine reliability prior: perception and measurement configurations.}
\label{tab:reliability_matrix_fine_perceive}

\begin{tabular*}{\textwidth}{@{\extracolsep{\fill}}lll*{7}{c}@{}}
\toprule
\multicolumn{10}{@{}l}{(a) Perception configurations} \\
\midrule
Mode & Backend & View &
\makecell{Image\\Caption} &
\makecell{Object\\Counting} &
\makecell{Object\\Presence} &
\makecell{Spatial\\Relationship} &
\makecell{Attribute\\Comparison} &
\makecell{Change\\Detection} &
\makecell{Scene\\Classification} \\
\midrule
\texttt{evidence} & \texttt{auto} & \texttt{auto}
& 0.900 & 0.958 & -- & 0.983 & 0.964 & 0.957 & -- \\
\texttt{evidence} & \texttt{auto} & \texttt{overview}
& 0.917 & -- & -- & -- & -- & -- & -- \\
\texttt{evidence} & \texttt{auto} & \texttt{tiles}
& -- & 0.909 & -- & 0.929 & -- & -- & -- \\
\texttt{evidence} & \texttt{vlm} & \texttt{auto}
& -- & -- & -- & 0.900 & -- & -- & -- \\
\texttt{objects} & \texttt{auto} & \texttt{auto}
& -- & 0.918 & 0.800 & 0.889 & 0.905 & 0.857 & -- \\
\texttt{objects} & \texttt{local} & \texttt{auto}
& -- & -- & -- & 0.929 & -- & -- & -- \\
\texttt{scene} & \texttt{auto} & \texttt{auto}
& -- & -- & -- & -- & -- & -- & 0.955 \\
\texttt{change} & \texttt{auto} & \texttt{auto}
& -- & -- & -- & -- & -- & 0.321 & -- \\
\end{tabular*}

\par\smallskip

\begin{tabular*}{\textwidth}{@{\extracolsep{\fill}}lll*{8}{c}@{}}
\midrule
\multicolumn{11}{@{}l}{(b) Measurement configurations} \\
\midrule
Metric & Relation & Sources &
\makecell{Object\\Counting} &
\makecell{Object\\Presence} &
\makecell{Object\\Localization} &
\makecell{Change\\Detection} &
\makecell{Spatial\\Relationship} &
\makecell{Physical\\Property} &
\makecell{Disaster\\Discrimination} &
\makecell{Environmental\\Assessment} \\
\midrule
\texttt{area} & \texttt{auto} & \texttt{regions}
& 0.818 & -- & -- & -- & 0.846 & 0.706 & 0.917 & -- \\
\texttt{count} & \texttt{auto} & \texttt{objects}
& 0.975 & 0.867 & 0.923 & 0.917 & -- & -- & 0.824 & -- \\
\texttt{count} &
\makecell[l]{\texttt{within\_}\\\texttt{distance}} &
\makecell[l]{\texttt{objects+}\\\texttt{objects}}
& -- & -- & -- & -- & 0.941 & -- & -- & -- \\
\texttt{distance} & \texttt{nearest} & \texttt{objects}
& -- & -- & -- & -- & 0.867 & -- & -- & -- \\
\texttt{distance} & \texttt{nearest} &
\makecell[l]{\texttt{objects+}\\\texttt{objects}}
& -- & -- & -- & -- & 0.846 & -- & -- & -- \\
\bottomrule
\end{tabular*}

\end{table*}

The full configuration matrix and the task label mapping will be released with the code.
\clearpage
\clearpage
\raggedbottom
\bibliographystyle{elsarticle-harv}
\bibliography{reference}
\end{document}